\documentclass[11pt]{article}

\usepackage[final]{acl}

\usepackage{times}
\usepackage{latexsym}
\usepackage{fvextra}
\usepackage[T1]{fontenc}
\usepackage[utf8]{inputenc}

\usepackage{microtype}
\usepackage{xcolor}

\newcommand{\add}[1]{\textcolor{black}{#1}}

\newenvironment{added}{\color{black}}{}   

\usepackage{inconsolata}

\usepackage{graphicx}
\usepackage{booktabs}
\usepackage{graphicx}

\usepackage{xcolor}
\usepackage{listings}
\usepackage{fancyhdr}
\usepackage{multirow}
\usepackage{amsmath} 
\usepackage{courier} % To set Courier as the font for listings
\usepackage{pdfpages}
\usepackage{adjustbox}
\usepackage{float}
\usepackage{lscape}

\usepackage{pifont}

\usepackage{amsmath}
\usepackage{bbm}
\usepackage{bm}

\usepackage{makecell}
\usepackage{wrapfig}
\usepackage{longtable}

\usepackage{float} % 导入float宏包

\definecolor{codebg}{rgb}{0.95,0.95,0.95}
\definecolor{keywords}{rgb}{0,0,1}    % 关键字颜色
\definecolor{comment}{rgb}{0.5,0.5,0.5} % 注释颜色
\definecolor{string}{rgb}{0.58,0.0,0.82} % 字符串颜色

\usepackage[most]{tcolorbox}

\definecolor{assistantonecolor}{RGB}{19,118,188}
\definecolor{assistanttwocolor}{RGB}{229,91,43}

\newcommand{\assistanttwomsg}[1]{\textcolor{assistanttwocolor}{{\textbf{#1: }}}}

\usepackage{verbatim}  % 加在导言区

\usepackage{xcolor}
\usepackage{listings}
\tcbset{
  aiboxbreakable/.style={
    width=\columnwidth, 
    top=10pt,
    colback=black!05,
    colframe=black!20,
    colbacktitle=black!50,
    enhanced,
    center,
    breakable,
    attach boxed title to top left={yshift=-0.1in,xshift=0.15in},
    boxed title style={boxrule=0pt,colframe=white,},
  }
}
\newtcolorbox{AIBoxBreak}[2][]{aiboxbreakable,title=#2,#1}

\title{A Skill-augmented Agentic Framework and Benchmark \\for Multi-Video Understanding} 

\author{
Yue Zhang\thanks{Equal contribution.} \quad
Liqiang Jing\footnotemark[1] \quad
Jia Li \quad
Yapeng Tian \quad
Xinya Du \quad
Yunhui Guo \quad
Vibhav Giridhar Gogate \\
Department of Computer Science \\
The University of Texas at Dallas, Richardson, TX, USA \\
\texttt{\{yue.zhang, liqiang.jing, vibhav.gogate\}@utdallas.edu}
}
\begin{document}
\maketitle
\begin{abstract}
Multimodal Large Language Models have achieved strong performance in single-video understanding, yet their ability to reason across multiple videos remains limited. Existing approaches typically concatenate multiple videos into a single input and perform direct inference, which introduces training–inference mismatch, information loss from frame compression, and a lack of explicit cross-video coordination. Meanwhile, current multi-video benchmarks primarily emphasize event-level comparison, leaving identity-level matching, fine-grained discrimination, and structured multi-step reasoning underexplored. To address these gaps, we introduce MVX-Bench, a Multi-Video Cross-Dimension Benchmark that reformulates 11 classical computer vision tasks into a unified multi-video question-answering framework, comprising 1,442 questions over 4,255 videos from diverse real-world datasets. We further propose SAMA, a Skill-Augmented Agentic Framework for Multi-Video Understanding, which integrates visual tools, task-specific skills, and a conflict-aware verification mechanism to support iterative and structured reasoning. Experiments show that SAMA improves substantially over its visual perception backbone and achieves the best average accuracy on MVX-Bench. Ablations confirm the benefits of skill-guided orchestration and conflict-aware verification.
\end{abstract}
\section{Introduction}
Multimodal Large Language Models (MLLMs)  integrate visual and textual inputs, enabling a comprehensive understanding of spatial relationships, object interactions, scene context, and abstract concepts. Early research primarily focused on image-based tasks such as image captioning and visual question answering (VQA), where reasoning is performed over static visual content. However, many real-world scenarios involve dynamic visual streams rather than isolated images, motivating a transition toward video understanding tasks, including video captioning and temporal question answering. While MLLMs have achieved substantial progress in single-video reasoning, their ability to model spatiotemporal patterns across multiple videos remains limited, as multi-video understanding requires cross-video alignment, consistency management, and structured reasoning beyond single-stream analysis.

\begin{figure}[t]
\centering
\includegraphics[width=0.49\textwidth]{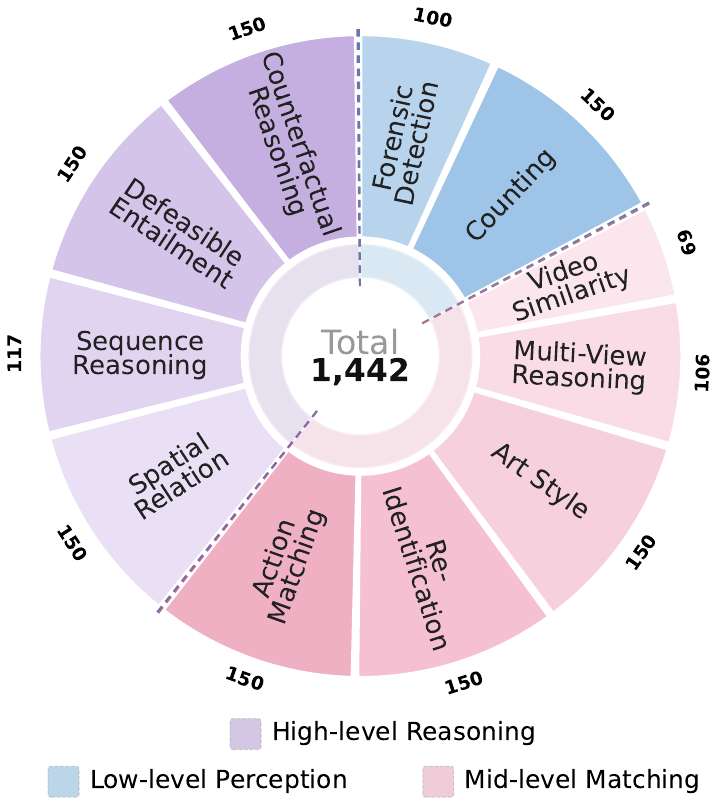}
\caption{Statistics of our benchmark. The benchmark includes 1,442 samples across 11 tasks, spanning low-level perception, mid-level cross-video matching, and high-level reasoning.}
\label{fig:dataset_stats}
\end{figure}

To facilitate systematic evaluation of multi-video understanding, recent benchmarks such as CrossVid~\cite{Li2025CrossVidAC}, MVU-Eval~\cite{Peng2025MVUEvalTM}, and CVBench~\cite{zhu2026cvbenchbenchmarkingcrossvideosynergies} have been introduced to evaluate cross-video reasoning capabilities of multimodal LLMs. These works provide valuable testbeds for assessing event-level understanding and comparative reasoning across multiple video streams. 
However, a substantial portion of existing multi-video datasets rely on synthetic or model-generated question–answer pairs, which may introduce distributional biases and reduce the diversity of real-world visual complexity. Moreover, current evaluations primarily emphasize event-level correspondence or comparative reasoning, leaving identity-level matching, fine-grained discrimination, and structurally complex multi-step reasoning comparatively underexplored. As a result, the current evaluation landscape may not fully reflect the breadth and difficulty of real-world multi-video understanding.

To address these limitations and provide a more comprehensive and realistic evaluation framework for multi-video understanding, we introduce MVX-Bench, a Multi-Video Cross-Dimension Benchmark. 
MVX-Bench reimagines classical computer vision problems within a unified multi-video question-answering paradigm, enabling systematic evaluation of MLLMs under diverse and realistic settings. As illustrated in Figure~\ref{fig:dataset_stats}, MVX-Bench consists of 11 canonical computer vision tasks adapted to multi-video scenarios, spanning low-level pattern matching (e.g., Counting), mid-level reasoning (e.g., Action Matching), and high-level visual understanding (e.g., structured cross-video reasoning and counterfactual inference).
Each traditional task is reformulated into a modern multiple-choice question-answering format, where answer options may consist of either videos or text. In total, MVX-Bench contains 1,442 questions across 4,255 videos. Each question involves multiple videos curated from a diverse range of datasets \cite{AminiNaieni2025OpenWorldOC, Zhang2025CanVL, Zi2020WildDeepfakeAC}, covering indoor household scenes as well as outdoor urban and natural environments. The questions and answer choices are either derived from the original datasets or manually authored to ensure clarity and controlled evaluation.

Existing multimodal approaches typically handle multi-video tasks by concatenating multiple videos into a single input sequence and performing direct inference. However, this single-pass paradigm often suffers from a mismatch between single-video training and multi-video inference, information loss caused by aggressive frame compression, and the absence of explicit mechanisms for coordinating reasoning across multiple video streams.
% struggles with distribution mismatch, information compression, and the lack of explicit mechanisms for structured cross-video coordination. 
Motivated by these limitations and inspired by recent video agent frameworks \cite{fan2024videoagent}, we propose SAMA, a Skill-Augmented Agentic Framework for Multi-Video Understanding, as illustrated in Figure~\ref{fig:method_overview}. 
Unlike existing video agents primarily designed for single-video reasoning, SAMA is among the first agentic frameworks explicitly tailored to multi-video scenarios. It incorporates a diverse set of visual tools and task-specific agentic skills designed for cross-video coordination, and further introduces a conflict resolution mechanism to address inconsistencies arising from heterogeneous tool outputs.
Experimental results on MVX-Bench demonstrate the effectiveness of our agentic framework, which outperforms all evaluated baselines, including open-source model families such as InternVL3.5 \cite{Wang2025InternVL35AO}, Qwen2.5-VL \cite{Bai2025Qwen25VLTR}, and Gemma-3 \cite{Kamath2025Gemma3T}, as well as proprietary models such as GPT.

Our contributions can be summarized as follows:
1) We construct MVX-Bench, a Multi-Video Cross-Dimension Benchmark comprising 11 computer vision tasks adapted to multi-video scenarios, covering low-level pattern matching, mid-level reasoning, and high-level visual understanding.
2) To the best of our knowledge, we present the first agentic framework explicitly tailored for multi-video understanding across diverse perception, matching, and reasoning tasks. Our framework, SAMA, integrates diverse visual tools, task-specific agentic skills, and a conflict resolution mechanism to support structured cross-video reasoning.
3) Extensive experimental results demonstrate the effectiveness of SAMA, outperforming strong open-source baselines and GPT-4o on MVX-Bench. Ablation studies further validate the importance of our skill design and conflict detection mechanism.

\section{Related Work}
\label{sec:related}

\subsection{Multimodal Large Language Models}

Multimodal Large Language Models (MLLMs) have progressed from image understanding~\cite{Liu2023VisualIT,Liu2023ImprovedBW,Zhu2023MiniGPT4EV} to video understanding~\cite{Zhang2023VideoLLaMAAI,Li2023VideoChatCV,Maaz2023VideoChatGPTTD}. Recent models, such as Qwen2.5-VL~\cite{Bai2025Qwen25VLTR} and InternVL3.5~\cite{Wang2025InternVL35AO}, extend video understanding to longer temporal contexts. However, current MLLMs are still primarily developed for single-video inputs. When applied to multi-video settings, they typically concatenate frames from different sources into a single sequence, which can lead to training--inference mismatch, information loss under limited context budgets, and the absence of explicit mechanisms for cross-video coordination.

\subsection{Video Understanding Benchmarks}

Existing video benchmarks mainly focus on single-video understanding~\cite{Li2023MVBenchAC,Fu2024VideoMMETF,Zhou2024MLVUBM,Wu2024LongVideoBenchAB}. More recent benchmarks, including CVBench~\cite{zhu2026cvbenchbenchmarkingcrossvideosynergies}, CrossVid~\cite{Li2025CrossVidAC}, and MVU-Eval~\cite{Peng2025MVUEvalTM}, begin to examine cross-video reasoning. Other recent efforts further study specific aspects of multi-video understanding, including fine-grained video differences, procedural step differences, and semantic video similarity~\cite{DBLP:conf/iclr/BurgessWZRLDDY25, Nagarajan_2024_CVPR, DBLP:conf/iccv/QiuSYEMFHS25, DBLP:conf/nips/LiberatoriCVWRR25}. However, existing benchmarks typically focus on specific aspects of cross-video understanding, while comprehensive evaluation spanning identity-level matching, forensic discrimination, and more structured forms of reasoning remains less explored. To study these settings, we introduce MVX-Bench, which consists of 11 tasks organized into three cognitive levels.

\subsection{Agentic Frameworks for Video Understanding}

Agentic approaches augment language models with external tools for visual reasoning~\cite{Wang2024VideoAgentLV, Yang2024VCAVC, Zhi2025VideoAgent2ET, Suris2023ViperGPTVI}. 
These methods provide tool access but largely rely on the LLM to implicitly discover effective orchestration strategies. 
In complex settings, however, effective tool use often requires more explicit procedural guidance, such as when to invoke a tool, how to combine intermediate evidence, and when to stop. 
Recent work beyond video understanding~\cite{Li2026SkillsBenchBH} explores this idea through \emph{skills}, which provide structured guidance for tool use and reasoning.
However, existing video agents are mainly designed for single-video settings and do not explicitly support multi-video coordination.
To address this limitation, we propose SAMA, a multi-video agentic framework that integrates task-specific skills with an external verification layer for conflict detection and refinement.
\section{MVX-Bench}
\label{sec:benchmark}
Our goal is to comprehensively evaluate the multi-video understanding capabilities of existing Multimodal LLMs. 
While recent benchmarks have made important progress in evaluating cross-video perception, comparison, and temporal reasoning, most existing efforts primarily focus on event-level association, aggregation, or alignment across videos.
% Based on the observation that existing benchmarks predominantly focus on evaluating single-video understanding abilities or cross-video perception and comparative reasoning, w
We introduce MVX-Bench (Multi-Video Cross-Dimension Benchmark), a novel benchmark designed to extend the multi-video evaluation space to identity-level matching, style-level abstraction, manipulation discrimination, and structurally complex reasoning scenarios such as counterfactual inference and defeasible entailment.

\begin{figure*}[ht!]
\centering
\includegraphics[width=\textwidth]{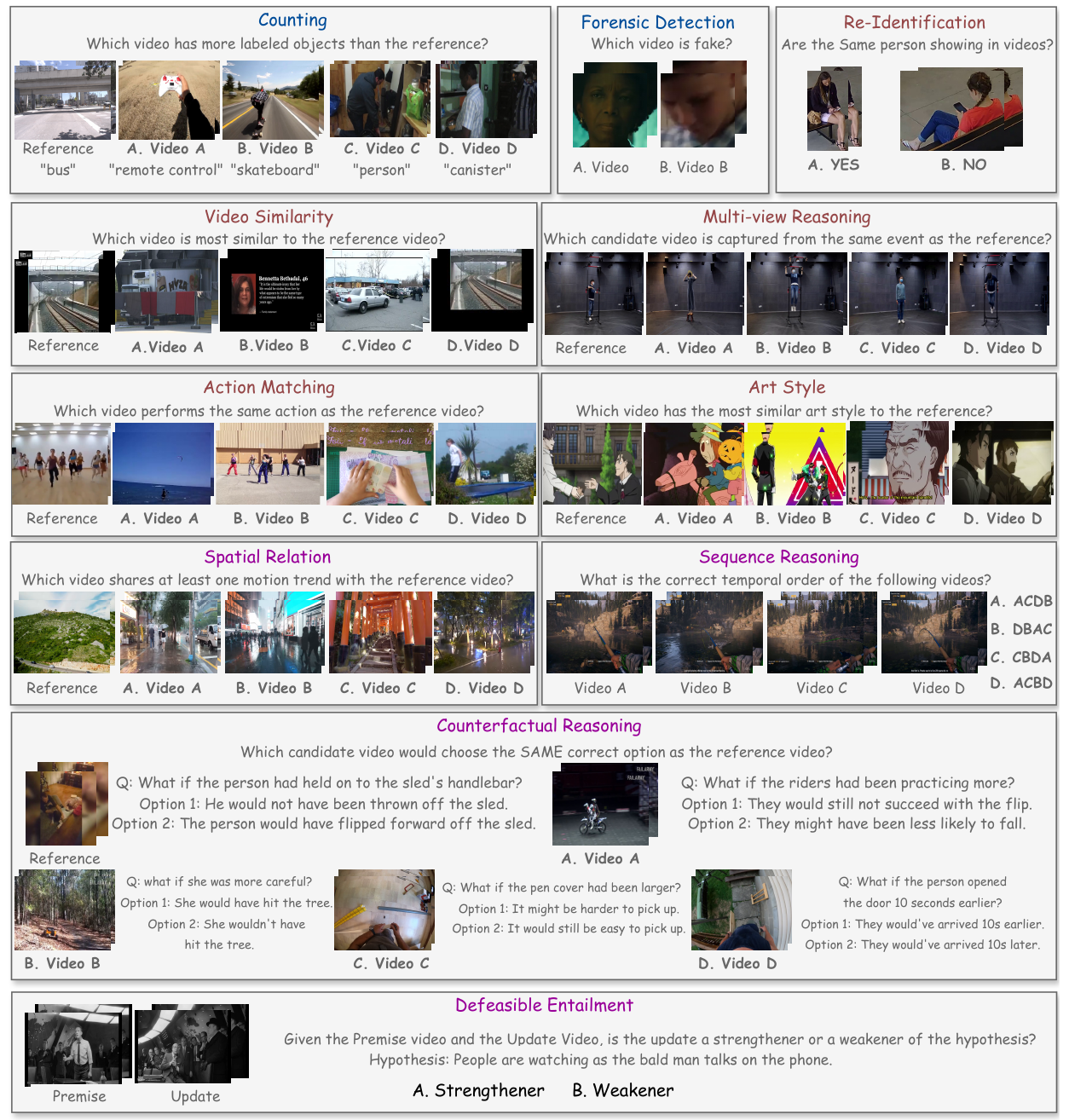}
\caption{\textbf{Overview of MVX-Bench.}
MVX-Bench evaluates 11 complementary capabilities spanning low-, mid-, and high-level video understanding and reasoning. Representative examples illustrate tasks ranging from perceptual comparison and cross-video matching to temporal, spatial, counterfactual, and defeasible reasoning.}
\label{fig:data}
\end{figure*}

\subsection{Overview}
\label{subsec:benchmark_overview}
%To systematically evaluate multi-video understanding, we carefully select 11 tasks that are difficult to solve.
%The tasks are drawn from both classic computer vision problems and real-world applications of Multimodal LLMs that inherently require multi-video understanding. They span a broad spectrum of capabilities, ranging from low-level perceptual analysis, to mid-level cross-video association and up to high-level complex reasoning.
%This diversity enables a systematic examination of Multimodal LLMs’ performance across different levels of abstraction and cognitive complexity in multi-video settings.

To rigorously evaluate multi-video understanding, we curated 11 challenging tasks (see Figure~\ref{fig:data}) representative of both classic computer vision problems and emerging real-world multimodal applications. Crucially, these tasks inherently demand cross-video reasoning. The benchmark spans a broad spectrum of capabilities, ranging from low-level perceptual analysis and mid-level cross-video association to high-level complex reasoning. This structured diversity enables a detailed examination of Multimodal LLM performance across varying levels of abstraction and cognitive complexity.

% To facilitate the evaluation, we recast all tasks into a unified multiple-choice question-answering format. We repurpose several existing vision datasets and additionally collect new data to ensure coverage of diverse multi-video scenarios.
% In total, our benchmark comprises 1,442 samples spanning 4,255 videos. Each sample consists of a question paired with multiple videos, requiring the model to integrate information across video inputs and select the correct answer from a set of candidate options.
% The questions and answer choices are either directly adapted from the original datasets or constructed using manually designed templates.
% Numbers of each task are reported in Figure \ref{fig:dataset_stats}.

To facilitate evaluation, we recast all tasks into a unified multiple-choice question-answering format. We build on annotated datasets to avoid model-generated labels and ensure controlled answer validity across diverse multi-video scenarios. In total, our benchmark comprises 1,442 samples across 4,255 videos. Each sample consists of a question paired with multiple videos, requiring the model to synthesize cross-video evidence to select the correct candidate option. The questions and answer choices are either adapted directly from the source datasets or generated via curated templates. Task-specific statistics are detailed in Figure \ref{fig:dataset_stats}.

\paragraph{Key features of MVX-Bench:} Unlike prior multi-video benchmarks that focus primarily on cross-video perception and comparative reasoning, our benchmark introduces identity-level matching, style-level abstraction, forensic discrimination, and logically structured cross-video reasoning (counterfactual \& defeasible inference), providing a qualitatively different stress-test for multi-video understanding.
First, prior works primarily focus on event-level or comparative reasoning across videos, whereas our benchmark introduces identity-level and style-level matching tasks (e.g., person re-identification and art style recognition), requiring models to disentangle invariant properties from superficial changes.
Second, we incorporate forensic detection and near-duplicate discrimination tasks, targeting subtle visual artifacts and fine-grained visual similarity—capabilities largely underexplored in existing multi-video QA benchmarks.
Third, beyond single-step comparison or aggregation, our benchmark introduces structurally complex reasoning scenarios, including nested counterfactual inference and defeasible entailment with belief revision. These tasks require multi-step reasoning across video evidence rather than simple cross-video alignment.

\subsection{Data Collection}

MVX-Bench is constructed by repurposing 11 existing video datasets into a unified multiple-choice evaluation format. Each task leverages dataset-provided annotations to create controlled cross-video reasoning problems, ensuring exactly one correct answer per question. Detailed construction procedures are provided in Appendix~\ref{appendix:data_construction_detailes}.

% MVX-Bench is constructed by repurposing existing annotated video datasets into a unified multiple-choice evaluation framework. 
% Each task leverages structured annotations (e.g., object counts, action labels, identity tags, similarity scores, temporal orderings, or entailment labels) to create controlled cross-video reasoning problems with exactly one correct answer. 
% Detailed dataset sources and construction procedures are provided in Appendix~\ref{appendix:data_construction_detailes}.

Specifically, (1) \textbf{Counting} evaluates quantitative comparison of object occurrences across videos; 
(2) \textbf{Forensic Detection} tests the ability to distinguish manipulated from authentic content; 
(3) \textbf{Action Matching} measures semantic alignment of actions across videos; 
(4) \textbf{Re-Identification} evaluates identity consistency under viewpoint and appearance variation; 
(5) \textbf{Art Style} assesses stylistic abstraction independent of depicted content; 
(6) \textbf{Multi-View Reasoning} probes spatial consistency across different viewpoints and sensing modalities; 
(7) \textbf{Video Similarity} measures fine-grained perceptual similarity; 
(8) \textbf{Counterfactual Reasoning} evaluates alignment of hypothetical outcomes across videos; 
(9) \textbf{Defeasible Entailment} tests dynamic belief revision under additional evidence; 
(10) \textbf{Sequence Reasoning} assesses temporal ordering capabilities; and 
(11) \textbf{Spatial Relation} evaluates understanding of motion dynamics and spatial trends.

\section{SAMA}
\label{sec:method}

Multimodal LLMs (MLLMs) have demonstrated strong performance on a wide range of video understanding tasks, including action recognition, temporal reasoning, and video question answering. Recent advances enable these models to process longer videos and perform increasingly complex visual-language reasoning.
However, extending these models from single-video understanding to multi-video scenarios remains non-trivial. A common approach is to concatenate multiple videos into a single input sequence and perform direct inference. While straightforward, this paradigm presents several structural challenges.
First, most MLLMs are predominantly trained on single-image or single-video data, with limited exposure to coordinated multi-video reasoning. As a result, direct multi-video inference introduces a distribution shift that these models are not explicitly optimized for.
Second, multi-video inputs significantly increase the number of visual tokens, often requiring aggressive frame sampling or compression, which may lead to information loss and reduced temporal fidelity.
Third, many multi-video tasks inherently require structured decomposition, such as per-video analysis, cross-video alignment, and multi-step reasoning, yet single-pass inference lacks explicit mechanisms for such modular coordination. 

While agentic frameworks have recently emerged to handle complex video tasks, existing approaches often fall short in multi-video scenarios due to two primary limitations. First, despite the availability of diverse information extraction tools, current video agents typically rely on a homogeneous and restricted toolset—often limited to basic captioning or tagging. This lack of tool diversity prevents the agent from capturing the multi-faceted evidence necessary to resolve cross-video contradictions or alignments. Second, these frameworks typically assume the underlying LLM can inherently orchestrate tools without explicit guidance, creating a procedural "how-to" gap. Without a mechanism to teach the model the specific logic of tool coordination, the agent struggles to effectively synthesize information from multiple streams. To bridge these gaps, we propose SAMA, a Skill-Augmented Agentic Framework for Multi-Video Understanding, which introduces a dedicated Skill Library to empower the model with both a rich set of specialized tools and the procedural mastery required for complex multi-video reasoning.

% To address these limitations, we propose SAMA, a Skill-Augmented Agentic Framework for Multi-Video Understanding.

% We proposed SAMA, a Skill-AugMented Agentic framework for multi-video understanding.

\subsection{Overview}
\label{subsec:overview}

\begin{figure*}[t]
\centering
\includegraphics[width=\textwidth]{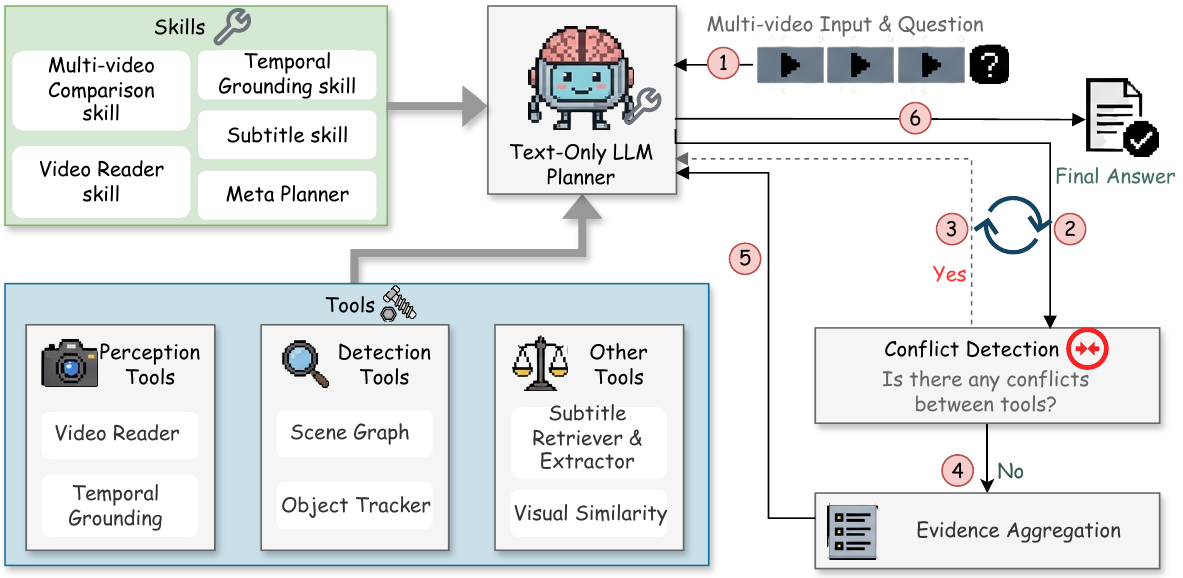}
\caption{Overview of the SAMA framework. A text-only LLM planner, guided by task-adaptive \textbf{skills} that define when, how, and in what order to invoke tools, orchestrates six visual tools across three groups: Perception, Detection, and Other tools. Skills determine the invocation strategy based on task type — for example, similarity tasks prioritize Visual Similarity before Video Reader, while counting tasks consult both Video Reader and Scene Graph to enable cross-modal conflict detection. Detected conflicts trigger adaptive re-reading, feeding corrective information back to the planner before the final answer is produced.}
\label{fig:method_overview}
\end{figure*}

As illustrated in Figure~\ref{fig:method_overview}, SAMA is organized into two layers: an Agent Core and an external Verification Layer.
The Agent Core consists of (i) a text-only LLM Planner that performs task analysis and decision-making, (ii) a multimodal toolkit of Tools for extracting evidence from videos, and (iii) reusable Skills that provide structured workflows and domain guidance to support tool use.
The Verification Layer includes a Conflict Detection module that identifies inconsistencies among tool outputs, an Evidence Aggregation module that consolidates verified information into an evidence pool, and an Adaptive Reading module that is triggered when conflicts are detected.
Upon detecting conflicting evidence, Adaptive Reading guides the Planner to invoke additional tools for further verification until the conflict is resolved.
After conflict resolution and aggregation, the Planner evaluates whether the accumulated evidence is sufficient to answer the question. If not, it initiates a new reasoning iteration by selecting additional tools or skills. Newly retrieved evidence is again processed through the Conflict Detection module before being incorporated into the evidence pool.

\subsection{Agent Core}
% \label{subsec:tools}

\subsubsection{Planner}
Large Language Models (LLMs) have demonstrated strong performance in multi-step reasoning, decision-making, and tool-augmented problem solving. These capabilities make them suitable for coordinating complex multi-video tasks that require planning and iterative reasoning. Following VideoAgent \cite{fan2024videoagent}, we adopt GPT-5.2~\cite{singh2026openaigpt5card} and Deepseek-V4-Pro~\cite{deepseekai2026deepseekv4} as text-only Planners. The Planner analyzes the question, selects appropriate tools and skills, generates queries, reasons over structured tool evidence, and determines when sufficient information has been collected to produce the final answer.
\subsubsection{Tools}

As shown in Figure \ref{fig:method_overview}, SAMA provides the planner with six tools organized into three functional groups, each offering a distinct mode of interaction with video content.

\noindent\textbf{Perception Tools:} 
% 1) Video Reader 读取视频 用自然语言来描述
% 1) \textit{Video Reader}, implemented using a large vision-language model (LVLM), answers the Planner's queries over video inputs in natural language.
% % the primary perception tool. Given a video identifier, a time interval, and a natural language question, it samples frames and queries a VLM to return a structured description: scene caption, actions, spatial layout, and question-relevant evidence. The frame rate adapts automatically --- coarse mode for initial overviews, fine mode for targeted follow-ups --- with a frame cache to avoid redundant extraction. 
% 2) \textit{Temporal Grounding}, which, given a query generated by the Planner, localizes the corresponding event or action within the video and returns the relevant timestamps.
% % This facilitates precise temporal localization for downstream reasoning.
1) \textit{Video Reader}, implemented with an LVLM, answers Planner queries over video inputs in natural language.
2) \textit{Temporal Grounding}, which localizes queried events or actions in the video and returns their timestamps.

\noindent\textbf{Detection Tools:} 1) \textit{Scene Graph}, which constructs a structured representation for each frame by extracting object categories, bounding boxes, object attributes, and relationships between objects.
% for open-vocabulary object detection, returning bounding boxes, confidence scores, HSV-derived colors, and relative positions. 
2) \textit{Object Tracker}, which associates detected objects across frames to preserve consistent identities, allowing temporal tracking of object counts, spatial trajectories, and attribute variations.
% 2) \textit{The Object Tracker} maintains cross-frame object identities to track count and attribute changes. 
% \textit{The Scene Detector} segments videos into coherent shots via frame histogram differences.

\noindent\textbf{Other Tools:} 
1) \textit{Visual Similarity}, which computes the similarity between frame-level embeddings of two video segments and returns a normalized similarity score in $[0,1]$.
% 2) \textit{The Subtitle Retriever \& Extractor} performs a semantic search over subtitle text, and the Subtitle Extractor returns raw subtitle content for a given time range.
2) \textit{Subtitle Retriever \& Extractor}, which performs semantic search over subtitle transcripts to retrieve relevant segments and extracts the corresponding raw subtitle text within a specified time range.

% Each tool accepts structured inputs and produces structured textual outputs. Crucially, these tools alone are insufficient --- without guidance on when and how to invoke them, an LLM planner struggles to use them effectively. This is where skills come in.

\subsubsection{Skills}
\label{subsec:skills}
Equipping an LLM planner with tool access alone is insufficient for multi-video tasks, as the planner must determine when to invoke each tool, in what order, and when to stop. To address this, we introduce task-specific skills, structured behavioral guides that wrap one or more tools with explicit invocation policies. 
%  Unlike standard tool descriptions that only specify functionality, a skill additionally encodes trigger conditions, invocation sequences, and stopping criteria.
Unlike generic prompt scaffolds, each skill is formulated as a reusable procedural module that encodes trigger conditions, invocation sequences, and stopping criteria for tool orchestration.
Each skill is provided in the Planner’s system context to guide structured reasoning, and SAMA includes five such skills, each targeting a distinct mode of multi-video reasoning. We provide details of all skills in Appendix \ref{sec:skill_prompt}.

% Equipping an LLM planner with tool access alone is insufficient for multi-video tasks, the planner must also know when to invoke each tool, in what order, and when to stop. We design a set of task-specific skills: structured behavioral guides that wrap one or more tools with explicit invocation policies. Unlike standard tool descriptions that only specify what a tool does, a skill additionally encodes when to invoke it, in what order relative to other tools, and when to stop gathering evidence. Each skill is provided to the planner as part of its system context and consists of trigger conditions that match question types to appropriate tools, an invocation protocol that prescribes the sequence of tool calls, and stopping criteria that define when evidence is sufficient to commit to an answer. SAMA includes five skills, each targeting a distinct mode of multi-video reasoning:

% \noindent\textbf{Multi-video Comparison:} This skill guides the planner through tasks that require comparing content across multiple videos. It encourages the planner to first establish a clear basis for comparison, maintain consistent queries across videos to ensure fair evidence gathering, and leverage the Similarity tool for direct visual comparison rather than relying on lengthy textual descriptions. The planner is advised to commit to an answer once evidence is decisive, rather than exhaustively examining all candidates.

\noindent\textbf{Multi-video Comparison:} 
This skill handles reference–candidate comparison tasks. The planner is instructed to first analyze the reference video, then evaluate candidate videos using consistent queries to ensure comparable evidence. For appearance-based questions, it prioritizes the \textit{Visual Similarity} tool for direct embedding-based comparison; for event-level matching, it relies on \textit{Video Reader} for semantic alignment and uses similarity scores as tie-breakers when necessary. The skill enforces evidence-based decision-making and commits to a final answer once decisive evidence is obtained.

\noindent\textbf{Video Reader:} This skill supports segment-level perception and evidence extraction. It requires the Planner to issue clear, self-contained questions targeting descriptive evidence, while reserving final judgment for the Planner rather than embedding decisions within the query. When initial evidence is insufficient, the skill prescribes sequential chunk-based scanning and targeted re-checking to refine the evidence before proceeding to reasoning.

% This skill guides general perception tasks. It instructs the planner to ask clear, self-contained questions targeting descriptive evidence, and to decide outside the tool rather than embedding judgment in the query. When information is insufficient, the skill directs sequential chunk scanning and re-checking rather than broad, unfocused queries. This prevents the common failure mode of posing vague questions that yield uninformative responses.

% \noindent\textbf{Temporal Grounding:} This skill defines a structured protocol for event localization in long videos. The Planner first invokes the Temporal Grounding tool to propose candidate timestamps based on entities and actions in the query. Retrieved timestamps are treated strictly as candidate hints rather than evidence and must be verified through the \textit{Video Reader} before being used in reasoning. If no relevant timestamps are returned, the skill enforces sequential chunk-based scanning to ensure comprehensive coverage.

\noindent\textbf{Temporal Grounding:} This skill defines a structured protocol for event localization in long videos. The Planner uses the Temporal Grounding tool to propose candidate timestamps from query entities and actions. These timestamps serve only as hints and must be verified by the \textit{Video Reader} before reasoning. If no relevant timestamps are found, the skill falls back to sequential chunk-based scanning.

% \noindent\textbf{Subtitle:} This skill defines a structured procedure for tasks involving spoken content, dialogue, or on-screen text. The Planner first invokes the \textit{Subtitle Retriever} to locate semantically relevant segments, followed by the \textit{Subtitle Extractor} to obtain exact transcript content within the identified time range. Retrieved text is treated as primary evidence, and visual verification via the \textit{Video Reader} is performed only when the textual content is ambiguous or insufficient for disambiguation.

\noindent\textbf{Subtitle:} This skill defines a structured procedure for tasks involving spoken content, dialogue, or on-screen text. The Planner uses the \textit{Subtitle Retriever} to find relevant segments and the \textit{Subtitle Extractor} to obtain exact transcripts. Retrieved text serves as primary evidence, while the \textit{Video Reader} is used only when textual evidence is ambiguous or insufficient.

% This skill is activated when questions involve spoken content, dialogue, or narration. It prescribes a two-stage strategy: semantic search first (Subtitle Retriever) to locate relevant passages, then precise extraction (Subtitle Extractor) for exact transcript content. Visual verification through the Video Reader is added only when the textual evidence is ambiguous.

\noindent\textbf{Meta Planner:} Unlike task-specific skills, this skill operates at a higher level to orchestrate multi-video reasoning. It requires the Planner to first identify the question type before invoking any specialized skill, ensuring appropriate skill selection and avoiding premature tool calls. The Meta Planner enforces consistent video-to-identifier mappings without assuming candidate ordering and resolves potential conflicts among multiple skills through a fixed priority scheme: output-format constraints take precedence, followed by task-specific rules, and finally default reasoning behaviors. 

% Unlike the preceding skills which target specific task types, this skill operates at a higher level: it guides the planner to first identify the nature of the question before engaging any task-specific skill, ensures that video-to-identifier mappings are respected without assuming ordering, and resolves conflicts when multiple skills provide competing guidance through a fixed priority — output-format constraints first, then task-specific constraints, then defaults.

% The central design principle across all skills is task-adaptive orchestration: different question types activate different skill configurations, leading to distinct tool usage patterns suited to each task's requirements.

\subsection{External Verification Layer}
\paragraph{Conflict Detection and Adaptive Re-reading.}
\label{subsec:conflict}
Even with well-designed skills, different tools may produce contradictory outputs. For example, the \textit{Video Reader}, powered by a VLM, may describe a traffic light as red, while the \textit{Scene Graph} tool's color analysis detects green in the majority of frames. Such conflicts arise because VLMs excel at high-level semantic interpretation but may be less reliable for fine-grained perceptual attributes such as color or object count.

To address this issue, SAMA leverages the Planner to automatically detect cross-tool inconsistencies by comparing structured outputs from different tools. When a conflict is identified, the system injects the conflicting evidence into the Planner’s context, prompting it to generate additional verification queries and invoke appropriate tools for further validation. The newly retrieved results are appended to the existing evidence pool, allowing the Planner to reassess and reconcile competing observations during reasoning. This verification loop continues until the Planner determines that the conflict has been sufficiently resolved, after which final decision-making proceeds based on the aggregated and validated evidence.

\paragraph{Evidence Aggregation.} 
After conflict resolution, verified evidence is aggregated before further reasoning. However, directly injecting all accumulated evidence into the Planner’s context may lead to a failure mode in which earlier observations are implicitly neglected as the context grows longer. 
To mitigate this issue, SAMA maintains a \textit{structured memory dictionary}, where each key corresponds to a video identifier and the associated value stores all validated evidence related to that video. For lengthy evidence, a concise, distilled representation is stored in memory to preserve essential information while controlling context length. 
After updating the memory, the Planner evaluates whether the accumulated evidence is sufficient to answer the question. If not, it initiates another reasoning cycle by calling tools or skills, retrieving new evidence, and routing it again through the conflict detection and aggregation process. This iterative loop continues until sufficient, validated evidence has been collected for the final answer.

% 在冲突解决后 进行证据aggregation 。 但是，在上下文中将所有的信息给出，会出现 failure mode where the planner ``forgets'' evidence from earlier rounds when the context grows long。 因此，为了解决这个问题，我们维护了一个memeory 词典 key是video value是和他相关的所有信息，同时对于过长的evidence只其其精简版本的信息存储到memory词典中。

% After evidence aggregation, the Planner evaluates whether the accumulated evidence is sufficient to answer the question. If not, it initiates another reasoning cycle by selecting additional tools or skills, retrieving new evidence, and passing it again through the conflict detection and aggregation process. 

% When the planner has gathered sufficient evidence, it constructs a structured summary before producing an answer. For multi-video understanding tasks, this takes the 

% form of a decision table: each candidate video is a row, the evidence dimensions are columns, and the planner reasons over this table to make its selection. This structured aggregation prevents a common failure mode where the planner ``forgets'' evidence from earlier rounds when the context grows long, and ensures that the final decision is grounded in an explicit, systematic comparison rather than implicit recollection.

\section{Experiments}
\label{sec:experiments}

\subsection{Experimental Setup}
We use accuracy as an evaluation metric for multi-video understanding tasks. We evaluate 19 open-source models spanning three model size tiers, including 1) Large ($\geq 14$B): InternVL3.5-38B, Qwen2.5-VL-32B-Instruct, InternVL3.5-14B, Gemma-3-27B-IT, Qwen3.6-27B~\cite{qwen3.6-27b}, and Qwen3.6-27B-Thinking; 2) Medium ($7$B $\leq$ Size $<14$B): Gemma-3-12B-IT, Pixtral-12B \cite{Agrawal2024Pixtral1}, InternVL3.5-8B, MiniCPM-V-2 (8B) \cite{Yao2024MiniCPMVAG}, Bee-8B~\cite{Zhang2025BeeAH}, Qwen2.5-VL-7B-Instruct, LLaVA-OneVision-7B~\cite{Li2024LLaVAOneVisionEV}, and Idefics3-8B-Llama3~\cite{Laurenon2024BuildingAB}; and 3) Small ($< 7$B): InternVL3.5-4B, Qwen2.5-VL-3B-Instruct, Gemma-3-4B-IT, Phi-3.5-Vision-Instruct (4.2B)~\cite{Abdin2024Phi3TR}, and Phi-4-Multimodal~\cite{Abouelenin2025Phi4MiniTR}. 
Our agentic framework employs GPT-5.2 as the text-only LLM Planner and Qwen2.5-VL-7B-Instruct as the Video Reader. Temporal grounding is implemented using LanguageBind~\cite{ICLR2024_2862ccf0}, while scene graph generation relies on GroundingDINO \cite{liu2023grounding}. Object tracking is performed with YOLOv8m~\cite{Varghese2024YOLOv8AN}, and visual similarity is computed using CLIP embeddings~\cite{Radford2021LearningTV}. Subtitle retrieval and extraction are supported by BGE-M3~\cite{Chen2024M3EmbeddingMM}. All results are reported with seed 42 and temperature 0.
The Planner is allowed to operate for up to 10 reasoning iterations per sample.

\subsection{Results and Analysis}

\begin{table*}[h!]
\centering
\adjustbox{max width=\textwidth}{
% \small
\begin{tabular}{lcccccccccccc}
\toprule
 & & \multicolumn{2}{c}{\textit{Low-Level}} & \multicolumn{5}{c}{\textit{Mid-Level}} & \multicolumn{4}{c}{\textit{High-Level}} \\
\textbf{Model} & \textbf{Avg} & \textbf{CNT} & \textbf{FD} & \textbf{AM} & \textbf{ART} & \textbf{RID} & \textbf{MVR} & \textbf{VS} & \textbf{CR} & \textbf{DE} & \textbf{SEQ} & \textbf{SPR} \\
\midrule
Random & 33.2\% & 25.0\% & 50.0\% & 25.0\% & 25.0\% & 50.0\% & 25.0\% & 25.0\% & 25.0\% & 50.0\% & 40.5\% & 25.0\% \\
GPT-4o & 47.7\% & 26.0\% & 49.0\% & 71.3\% & \underline{70.7\%} & 40.0\% & \textbf{70.8\%} & 76.8\% & 23.3\% & 52.0\% & 39.3\% & 26.7\% \\
GPT-5.2 & 50.9\% & 32.0\% & \textbf{68.0\%} & \textbf{79.3\%} & \textbf{73.3\%} & 58.0\% & 57.5\% & \underline{81.2\%} & 24.0\% & 52.7\% & 22.4\% & 28.7\% \\
\midrule
\multicolumn{13}{c}{\textit{Model Size} $\geq$ 14B} \\
\midrule
InternVL3.5-38B & 47.2\% & 28.7\% & 45.0\% & 71.3\% & 53.3\% & 58.0\% & 53.8\% & \textbf{82.6\%} & 22.7\% & 46.0\% & \underline{50.4\%} & 28.7\% \\
Qwen2.5-VL-32B & 46.3\% & 20.7\% & 52.0\% & 72.0\% & 50.7\% & 56.7\% & \underline{58.5\%} & 78.3\% & 19.3\% & 51.3\% & 45.3\% & 27.3\% \\
InternVL3.5-14B & 43.8\% & 26.7\% & 51.0\% & 68.0\% & 44.0\% & \textbf{62.0\%} & 33.0\% & 79.7\% & 22.7\% & 46.0\% & 40.2\% & 26.0\% \\
Gemma-3-27B-IT & 44.7\% & 28.0\% & 46.0\% & 72.7\% & 56.0\% & 56.7\% & 33.0\% & \textbf{82.6\%} & 16.7\% & 52.7\% & 35.9\% & 27.3\% \\
\add{Qwen3.6-27B}  & \add{34.7\%} & \add{28.7\%} & \add{57.0\%} & \add{30.7\%} & \add{24.0\%} & \add{57.3\%} & \add{35.8\%} & \add{26.1\%} & \add{20.0\%} & \add{47.3\%} & \add{40.2\%} & \add{19.3\%} \\
\add{Qwen3.6-27B-Thinking} & \add{45.9\%} & \add{\underline{44.7\%}} & \add{56.0\%} & \add{67.3\%} & \add{30.0\%} & \add{\underline{60.0\%}} & \add{\textbf{70.8\%}} & \add{53.6\%} & \add{16.0\%} & \add{47.3\%} & \add{44.4\%} & \add{\underline{29.3\%}} \\
\midrule
\multicolumn{13}{c}{\textit{Model Size}: 7B $\leq$ Size $<$ 14B} \\
\midrule
Gemma-3-12B-IT & 41.5\% & 25.3\% & 43.0\% & 65.3\% & 49.3\% & \underline{60.0\%} & 16.0\% & 71.0\% & 23.3\% & 52.0\% & 36.3\% & 22.0\% \\
Pixtral-12B & 35.4\% & 23.3\% & 43.0\% & 27.3\% & 33.3\% & 58.0\% & 27.4\% & 36.2\% & 30.7\% & 46.7\% & 39.3\% & 25.3\% \\
InternVL3.5-8B & 44.7\% & 30.7\% & 49.0\% & 68.7\% & 49.3\% & \underline{60.0\%} & 26.4\% & \underline{81.2\%} & 21.3\% & 51.3\% & 43.6\% & 26.0\% \\
MiniCPM-V-2 & 32.0\% & 28.7\% & 41.0\% & 24.0\% & 23.3\% & 51.3\% & 22.6\% & 15.9\% & 28.0\% & 49.3\% & 38.5\% & 22.0\% \\
Bee-8B & 33.2\% & 26.0\% & 55.0\% & 28.7\% & 23.3\% & 52.7\% & 34.9\% & 24.6\% & 26.7\% & 32.7\% & 45.1\% & 22.0\% \\
Qwen2.5-VL-7B & 32.6\% & 22.0\% & \underline{59.0\%} & 24.7\% & 21.3\% & 58.7\% & 21.7\% & 23.2\% & 22.0\% & 44.7\% & 44.4\% & 20.0\% \\
LLaVA-OneVision-7B & 32.7\% & 26.0\% & 47.0\% & 29.3\% & 26.0\% & 54.0\% & 22.6\% & 23.2\% & 18.0\% & 52.0\% & 40.5\% & 19.3\% \\
Idefics3-8B-Llama3 & 33.4\% & 25.3\% & 51.0\% & 31.3\% & 24.7\% & 54.7\% & 25.5\% & 27.5\% & 18.7\% & 50.7\% & 35.0\% & 23.3\% \\
\midrule
\multicolumn{13}{c}{\textit{Model Size} $<$ 7B} \\
\midrule
InternVL3.5-4B & 44.3\% & 26.7\% & 55.0\% & 64.7\% & 34.7\% & 57.3\% & 48.1\% & 79.7\% & 25.3\% & 53.3\% & 44.4\% & 22.0\% \\
Qwen2.5-VL-3B & 38.1\% & 29.3\% & 55.0\% & 45.3\% & 27.3\% & 56.0\% & 34.9\% & 63.8\% & 16.0\% & 47.3\% & 41.0\% & 22.7\% \\
Gemma-3-4B-IT & 40.6\% & 28.7\% & 49.0\% & 59.3\% & 30.7\% & 54.0\% & 16.0\% & 72.5\% & 26.7\% & \underline{56.0\%} & 42.7\% & 24.0\% \\
Phi-3.5-Vision & 22.8\% & 22.2\% & 13.0\% & 28.1\% & 8.8\% & 58.0\% & 19.8\% & 12.5\% & 18.8\% & 0.0\% & 9.1\% & 25.0\% \\
Phi-4-Multimodal & 34.4\% & 21.3\% & 55.0\% & 28.7\% & 23.3\% & 55.3\% & 34.9\% & 21.7\% & 27.3\% & 48.0\% & 42.6\% & 22.7\% \\
\midrule
\multicolumn{13}{c}{\textit{Our Method}} \\
\midrule
\add{\textbf{SAMA (Qwen2.5-VL-7B, DeepSeek-V4-Pro)}} & \add{\underline{51.7\%}} & \add{\textbf{49.3\%}} & \add{47.0\%} & \add{76.7\%} & \add{58.7\%} & \add{58.0\%} & \add{33.0\%} & \add{\textbf{82.6\%}} & \add{\textbf{42.0\%}} & \add{47.3\%} & \add{\textbf{59.0\%}} & \add{26.7\%} \\
\textbf{SAMA (Qwen2.5-VL-7B, \add {GPT-5.2})}  & \textbf{52.7\%} & 41.2\% & 53.3\% & \underline{78.0\%} & 64.6\% & 57.3\% & 54.0\% & 78.8\% & \underline{32.0\%} & \textbf{56.7\%} & 49.3\% & \textbf{30.0\%} \\
\bottomrule
\end{tabular}
}
\caption{\textbf{Overall results on our benchmark.} Tasks are grouped by category: Perception, Cross-Video Matching, and Reasoning. Best results are in \textbf{bold}; second-best are \underline{underlined}.}
\label{tab:main_results}
\end{table*}

Table~\ref{tab:main_results} presents the results across all 11 tasks. We highlight several key findings.
1) \textbf{Multi-video understanding remains far from solved.}
The best overall accuracy is 52.7\%, achieved by SAMA, and even GPT-5.2 achieves only 50.9\%. Most open-source models cluster between 32\% and 47\%, with several tasks near random baseline performance. This confirms that our benchmark poses a genuine challenge to current models.
2) \textbf{Scaling helps selectively, not uniformly.}
Larger models generally outperform smaller ones on association tasks such as Action Matching and Video Similarity, where recognizing visual patterns benefits from greater model capacity. However, scaling provides little advantage on reasoning tasks: on Counterfactual Reasoning, all models regardless of size score between 16\% and 32\%, barely above the 25\% random baseline. Similarly, Spatial Relation remains near random across all model scales. This suggests that certain multi-video reasoning capabilities are not emergent properties of scale.
3) \textbf{Task-specific strengths vary across models.} 
No single model dominates all tasks. GPT-5.2 leads on Action Matching and Forensic Detection, while InternVL3.5-38B achieves the best Sequence Reasoning and Video Similarity. Notably, InternVL3.5-14B achieves the highest Re-Identification accuracy, outperforming much larger models, suggesting that identity reasoning may depend more on training data composition than raw model size.
4) \textbf{Skill-augmented agentic reasoning substantially improves a 7B vision model.}
SAMA uses Qwen2.5-VL-7B as its Video Reader, while the same model achieves only 32.6\% when used end-to-end. With the agentic framework, the same Video Reader reaches 52.7\%, an improvement of \textbf{20.1 percentage points}. This gain is consistent across nearly all tasks.
% : Action Matching rises from 24.7\% to 78.0\%, Art Style from 21.3\% to 64.6\%, and Counting from 22.0\% to 41.2\%. The improvement is especially pronounced on tasks that require multi-step inference: Counterfactual Reasoning improves from 22.0\% to 32.0\%, and Spatial Relation from 20.0\% to 30.0\% --- tasks where all end-to-end models, regardless of size, remain near random. 
These results demonstrate that structured tool orchestration can unlock capabilities that are inaccessible to the same vision model when used in a standard end-to-end pipeline.
5) \textbf{SAMA reaches proprietary-model performance with a 7B Video Reader.}
Beyond outperforming its direct Video Reader baseline, SAMA achieves 52.7\% overall accuracy, comparable to GPT-5.2 and surpassing GPT-4o as well as all open-source end-to-end models including InternVL3.5-38B. SAMA shows particular strength on tasks requiring fine-grained perception and complex reasoning, improving Counting from 32.0\% to 41.2\%, Counterfactual Reasoning from 24.0\% to 32.0\%, Defeasible Entailment from 52.7\% to 56.7\%, and Sequence Reasoning from 22.4\% to 49.3\%.
% --- while GPT-5.2 retains advantages on association tasks like Action Matching and Art Style where its larger vision encoder provides stronger feature representations. This complementary pattern suggests that agentic frameworks and large-scale end-to-end models excel at fundamentally different aspects of multi-video understanding.
6) \textbf{SAMA generalizes to open-source planners.} 
With DeepSeek-V4-Pro as the planner, SAMA still attains 51.7\% overall accuracy, surpassing GPT-5.2 end-to-end and approaching the GPT-5.2-planner variant. This confirms that SAMA's gains stem from its skill-augmented architecture rather than the underlying planner.

\subsection{Ablation Studies}
To evaluate the contributions of the skills, conflict detection, and evidence aggregation modules, we randomly sample 350 examples and report the results in Table~\ref{tab:ablation}.
1) \textbf{Effect of skill design.}
We compare the SAMA with a variant that removes all skills, retaining only basic tool interfaces. Removing skills reduces accuracy from 60.0\% to 51.7\%, corresponding to an 8.3 percentage point drop. 
Without structured skill guidance, the Planner defaults to generic \textit{Video Reader} calls and underutilizes specialized tools such as \textit{Visual Similarity} and \textit{Scene Graph}, even when they are appropriate for the task. These results demonstrate that explicit skill design provides essential orchestration policies beyond standalone tool descriptions.
2) \textbf{Effect of conflict detection.}
We further ablate the Conflict Detection and Adaptive Reading modules (w/o-Conflict). Removing this verification mechanism leads to a noticeable performance decline, indicating that resolving cross-tool inconsistencies is critical for reliable multi-video reasoning. In particular, without conflict detection, contradictory tool outputs remain unverified, which propagates errors into the final decision stage. 
3) \textbf{Effect of evidence aggregation.}
We further ablate the Evidence Aggregation module (w/o-Evidence Aggregation). In this variant, verified observations are directly appended to the Planner context rather than organized into a structured per-video evidence memory. Removing evidence aggregation leads to a performance decline, indicating that structured evidence organization is beneficial for retaining and comparing information across multiple videos during reasoning.
\add{More ablations on the LLM planner and Video Reader, data quality analysis, cross-benchmark generalization, and runtime statistics are in Appendices~\ref{app:extended-ablation}, \ref{app:data-quality}, \ref{app:generalization}, and \ref{app:runtime}.}

\begin{table}[h]
\centering

\begin{tabular}{lc}
\toprule
\textbf{Configuration} & \textbf{Acc.} \\
\midrule
SAMA & 60.0\% \\ \midrule
\quad w/o-Skills & 51.7\% \\
\quad w/o-Conflict & 55.9\% \\
\quad w/o-Evidence Aggregation & 57.3\% \\
\bottomrule
\end{tabular}
\caption{Ablation studies of our SAMA.}
\label{tab:ablation}
\end{table}

\section{Conclusion}
In this work, we revisit the problem of multi-video understanding and argue that existing evaluation paradigms and inference strategies are insufficient for capturing the structural complexity of cross-video reasoning. We introduce MVX-Bench, a comprehensive benchmark that extends classical computer vision tasks into multi-video settings, enabling systematic evaluation across low-level matching, mid-level reasoning, and high-level structured inference. To address the limitations of direct multi-video concatenation, we propose SAMA, a skill-augmented agentic framework that integrates diverse visual tools, task-specific skills, and conflict-aware verification to support iterative and coordinated reasoning across video streams. Results show that skill-guided orchestration and conflict-aware verification improve multi-video reasoning, especially when evidence must be decomposed and reconciled across videos.
% improvements over strong baselines, highlighting the importance of structured orchestration and explicit evidence management. 
% We hope that MVX-Bench and SAMA provide a foundation for future research on robust and scalable multi-video reasoning in multimodal language models.

\section*{Limitations}
Our benchmark focuses on multiple-choice question-answering settings, which may not fully reflect open-ended generation scenarios. Exploring free-form multi-video reasoning remains an interesting direction for future work. Additionally, SAMA processes audio only through subtitle retrieval; integrating native audio tools (e.g., Whisper, AudioCLIP) as additional skills for cross-video synchronization is left for future work. Since MVX-Bench repurposes public video datasets, some source videos may overlap with the pretraining data of evaluated models, particularly for closed-source models whose training corpora cannot be audited. Although MVX-Bench constructs new cross-video pairings and QA instances rather than directly reusing the original tasks, we cannot fully rule out potential pretraining overlap.

\section*{Acknowledgements}
This work was funded in part by the DARPA ANSR and CODORD programs under Awards FA8750-23-2-0004 and HR00112590089, respectively, and by the National Science Foundation (NSF) under Grant Nos. 2513070 and 2612287. This work was also supported by the OpenAI Researcher Access Program under Award No. 0000013448.

\bibliography{custom}

@String(CVPR  = {IEEE Conf. Comput. Vis. Pattern Recog.})

@String(ICCV  = {Int. Conf. Comput. Vis.})

@String(ECCV  = {Eur. Conf. Comput. Vis.})

@String(NeurIPS = {Adv. Neural Inform. Process. Syst.})

@String(ICLR  = {Int. Conf. Learn. Represent.})

@String(AAAI  = {AAAI})

@String(ICASSP=	{ICASSP})

@String(CVPR  = {CVPR})

@String(ICCV  = {ICCV})

@String(ECCV  = {ECCV})

@String(NeurIPS = {NeurIPS})

@String(ICLR  = {ICLR})

@InProceedings{fan2024videoagent,
author="Fan, Yue
and Ma, Xiaojian
and Wu, Rujie
and Du, Yuntao
and Li, Jiaqi
and Gao, Zhi
and Li, Qing",
editor="Leonardis, Ale{\v{s}}
and Ricci, Elisa
and Roth, Stefan
and Russakovsky, Olga
and Sattler, Torsten
and Varol, G{\"u}l",
title="VideoAgent: A Memory-Augmented Multimodal Agent for Video Understanding",
booktitle="Computer Vision -- ECCV 2024",
year="2025",
publisher="Springer Nature Switzerland",
address="Cham",
pages="75--92",
isbn="978-3-031-72670-5"
}

@inproceedings{Liu2023VisualIT,
  author = {Haotian Liu and Chunyuan Li and Qingyang Wu and Yong Jae Lee},
  title = {Visual Instruction Tuning},
  booktitle = NeurIPS,
  year = {2023}
}

@inproceedings{DBLP:conf/wacv/DavilaDLFPCCBMHC23,
  author = {Daniel Davila and Dawei Du and Bryon Lewis and Christopher Funk and Joseph Van Pelt and Roderic Collins and Kellie Corona and Matt Brown and Scott McCloskey and Anthony Hoogs and Brian Clipp},
  title = {MEVID: Multi-view Extended Videos with Identities for Video Person Re-Identification},
  booktitle = {IEEE/CVF Winter Conference on Applications of Computer Vision (WACV)},
  year = {2023}
}

@inproceedings{Li2025CrossVidAC,
  author       = {Jingyao Li and
                  Jingyun Wang and
                  Molin Tan and
                  Haochen Wang and
                  Cilin Yan and
                  Likun Shi and
                  Jiayin Cai and
                  Xiaolong Jiang and
                  Yao Hu},
  editor       = {Sven Koenig and
                  Chad Jenkins and
                  Matthew E. Taylor},
  title        = {CrossVid: {A} Comprehensive Benchmark for Evaluating Cross-Video Reasoning
                  in Multimodal Large Language Models},
  booktitle    = {Fortieth {AAAI} Conference on Artificial Intelligence, Thirty-Eighth
                  Conference on Innovative Applications of Artificial Intelligence,
                  Sixteenth Symposium on Educational Advances in Artificial Intelligence,
                  {AAAI} 2026, Singapore, January 20-27, 2026},
  pages        = {6244--6252},
  publisher    = {{AAAI} Press},
  year         = {2026},
  url          = {https://doi.org/10.1609/aaai.v40i8.37550},
  doi          = {10.1609/AAAI.V40I8.37550},
  bibsource    = {dblp computer science bibliography, https://dblp.org}
}

@inproceedings{Peng2025MVUEvalTM,
 author = {Peng, Tianhao and Wang, Haochen and Zhang, Yuanxing and Wang, Noah and Wang, Zili and Zhang, Ge and Yang, Jian and Li, Shihao and Wang, Yanghai and Wang, Xintao and Li, Houyi and Ji, Wei and Wan, Pengfei and Huang, Wenhao and ZHANG, ZHAO-XIANG and Liu, Jiaheng},
 booktitle = {Advances in Neural Information Processing Systems},
 doi = {10.52202/085713-1800},
 editor = {D. Belgrave and C. Zhang and H. Lin and R. Pascanu and P. Koniusz and M. Ghassemi and N. Chen},
 pages = {},
 publisher = {Curran Associates, Inc.},
 title = {MVU-Eval: Towards Multi-Video Understanding Evaluation for Multimodal LLMs},
 url = {https://proceedings.neurips.cc/paper_files/paper/2025/file/4dbb159f2c68b359b3b6ed7dd039481c-Paper-Datasets_and_Benchmarks_Track.pdf},
 volume = {38, Main Conference},
 year = {2025}
}

@inproceedings{AminiNaieni2025OpenWorldOC,
  author       = {Niki Amini{-}Naieni and
                  Andrew Zisserman},
  editor       = {Sven Koenig and
                  Chad Jenkins and
                  Matthew E. Taylor},
  title        = {Open-World Object Counting in Videos},
  booktitle    = {Fortieth {AAAI} Conference on Artificial Intelligence, Thirty-Eighth
                  Conference on Innovative Applications of Artificial Intelligence,
                  Sixteenth Symposium on Educational Advances in Artificial Intelligence,
                  {AAAI} 2026, Singapore, January 20-27, 2026},
  pages        = {2300--2308},
  publisher    = {{AAAI} Press},
  year         = {2026},
  url          = {https://doi.org/10.1609/aaai.v40i4.37214},
  doi          = {10.1609/AAAI.V40I4.37214},
  bibsource    = {dblp computer science bibliography, https://dblp.org}
}

@inproceedings{Zi2020WildDeepfakeAC,
author = {Zi, Bojia and Chang, Minghao and Chen, Jingjing and Ma, Xingjun and Jiang, Yu-Gang},
title = {WildDeepfake: A Challenging Real-World Dataset for Deepfake Detection},
year = {2020},
isbn = {9781450379885},
publisher = {Association for Computing Machinery},
address = {New York, NY, USA},
url = {https://doi.org/10.1145/3394171.3413769},
doi = {10.1145/3394171.3413769},
booktitle = {Proceedings of the 28th ACM International Conference on Multimedia},
pages = {2382–2390},
numpages = {9},
location = {Seattle, WA, USA},
series = {MM '20}
}

@misc{Carreira2019ASN,
      title={A Short Note on the Kinetics-700 Human Action Dataset}, 
      author={Joao Carreira and Eric Noland and Chloe Hillier and Andrew Zisserman},
      year={2022},
      eprint={1907.06987},
      archivePrefix={arXiv},
      primaryClass={cs.CV},
      url={https://arxiv.org/abs/1907.06987}, 
}

@INPROCEEDINGS{Cai2024Anim400KAL,
  author={Cai, Kevin and Liu, Chonghua and Chan, David M.},
  booktitle={ICASSP 2024 - 2024 IEEE International Conference on Acoustics, Speech and Signal Processing (ICASSP)}, 
  title={Anim-400K: A Large-Scale Dataset for Automated End to End Dubbing of Video}, 
  year={2024},
  volume={},
  number={},
  pages={11796-11800},
  doi={10.1109/ICASSP48485.2024.10447911}}

@ARTICLE{10403933,
  author={Guo, Shuai and Hu, Jingchuan and Zhou, Kai and Wang, Jionghao and Song, Li and Xie, Rong and Zhang, Wenjun},
  journal={IEEE Transactions on Multimedia}, 
  title={Real-Time Free Viewpoint Video Synthesis System Based on DIBR and A Depth Estimation Network}, 
  year={2024},
  volume={},
  number={},
  pages={1-16},
  doi={10.1109/TMM.2024.3355639}
}

@article{KordopatisZilos2018FIVRFI,
author = {Kordopatis-Zilos, Giorgos and Papadopoulos, Symeon and Patras, Ioannis and Kompatsiaris, Ioannis},
title = {FIVR: Fine-Grained Incident Video Retrieval},
year = {2019},
issue_date = {Oct. 2019},
publisher = {IEEE Press},
volume = {21},
number = {10},
issn = {1520-9210},
url = {https://doi.org/10.1109/TMM.2019.2905741},
doi = {10.1109/TMM.2019.2905741},
journal = {Trans. Multi.},
month = oct,
pages = {2638–2652},
numpages = {15}
}

@inproceedings{Wu2023ACQUIREDAD,
    title = "{ACQUIRED}: A Dataset for Answering Counterfactual Questions In Real-Life Videos",
    author = "Wu, Te-Lin  and
      Dou, Zi-Yi  and
      Hu, Qingyuan  and
      Hou, Yu  and
      Chandra, Nischal  and
      Freedman, Marjorie  and
      Weischedel, Ralph  and
      Peng, Nanyun",
    editor = "Bouamor, Houda  and
      Pino, Juan  and
      Bali, Kalika",
    booktitle = "Proceedings of the 2023 Conference on Empirical Methods in Natural Language Processing",
    month = dec,
    year = "2023",
    address = "Singapore",
    publisher = "Association for Computational Linguistics",
    url = "https://aclanthology.org/2023.emnlp-main.719/",
    doi = "10.18653/v1/2023.emnlp-main.719",
    pages = "11753--11770"
}

@misc{Zhang2025CanVL,
      title={Can Video Large Multimodal Models Think Like Doubters-or Double-Down: A Study on Defeasible Video Entailment}, 
      author={Yue Zhang and Jilei Sun and Yunhui Guo and Vibhav Gogate},
      year={2025},
      eprint={2506.22385},
      archivePrefix={arXiv},
      primaryClass={cs.CV},
      url={https://arxiv.org/abs/2506.22385}, 
}

@InProceedings{Ju2025CIVIDAC,
    author    = {Ju, Yiming and Hu, Jijin and Luo, Zhengxiong and Deng, Haoge and Zhao, hanyu and Du, Li and Xiao, Wenbo and Wu, Chengwei and Hao, Donglin and Wang, Xinlong and Pan, Tengfei},
    title     = {CI-VID: A Coherent Interleaved Text-Video Dataset},
    booktitle = {Proceedings of the IEEE/CVF Conference on Computer Vision and Pattern Recognition (CVPR)},
    month     = {June},
    year      = {2026},
    pages     = {25568-25577}
}

@InProceedings{Wang2025SpatialVIDAL,
    author    = {Wang, Jiahao and Yuan, Yufeng and Zheng, Rujie and Lin, Youtian and Gao, Jian and Chen, Lin-Zhuo and Bao, Yajie and Zeng, Chang and Zhou, Yanxi and Long, Xiao-Xiao and Zhu, Hao and Zhang, Zhaoxiang and Cao, Xun and Yao, Yao},
    title     = {SpatialVID: A Large-Scale Video Dataset with Spatial Annotations},
    booktitle = {Proceedings of the IEEE/CVF Conference on Computer Vision and Pattern Recognition (CVPR)},
    month     = {June},
    year      = {2026},
    pages     = {42592-42603}
}

@misc{zhu2026cvbenchbenchmarkingcrossvideosynergies,
      title={CVBench: Benchmarking Cross-Video Synergies for Complex Multimodal Reasoning}, 
      author={Nannan Zhu and Yonghao Dong and Teng Wang and Xueqian Li and Shengjun Deng and Yijia Wang and Zheng Hong and Tiantian Geng and Guo Niu and Hanyan Huang and Xiongfei Yao and Shuaiwei Jiao},
      year={2025},
      eprint={2508.19542},
      archivePrefix={arXiv},
      primaryClass={cs.CV},
      url={https://arxiv.org/abs/2508.19542}, 
}

@misc{Wang2025InternVL35AO,
      title={InternVL3.5: Advancing Open-Source Multimodal Models in Versatility, Reasoning, and Efficiency}, 
      author={Weiyun Wang and Zhangwei Gao and Lixin Gu and Hengjun Pu and Long Cui and Xingguang Wei and Zhaoyang Liu and Linglin Jing and Shenglong Ye and Jie Shao and Zhaokai Wang and Zhe Chen and Hongjie Zhang and Ganlin Yang and Haomin Wang and Qi Wei and Jinhui Yin and Wenhao Li and Erfei Cui and Guanzhou Chen and Zichen Ding and Changyao Tian and Zhenyu Wu and Jingjing Xie and Zehao Li and Bowen Yang and Yuchen Duan and Xuehui Wang and Zhi Hou and Haoran Hao and Tianyi Zhang and Songze Li and Xiangyu Zhao and Haodong Duan and Nianchen Deng and Bin Fu and Yinan He and Yi Wang and Conghui He and Botian Shi and Junjun He and Yingtong Xiong and Han Lv and Lijun Wu and Wenqi Shao and Kaipeng Zhang and Huipeng Deng and Biqing Qi and Jiaye Ge and Qipeng Guo and Wenwei Zhang and Songyang Zhang and Maosong Cao and Junyao Lin and Kexian Tang and Jianfei Gao and Haian Huang and Yuzhe Gu and Chengqi Lyu and Huanze Tang and Rui Wang and Haijun Lv and Wanli Ouyang and Limin Wang and Min Dou and Xizhou Zhu and Tong Lu and Dahua Lin and Jifeng Dai and Weijie Su and Bowen Zhou and Kai Chen and Yu Qiao and Wenhai Wang and Gen Luo},
      year={2025},
      eprint={2508.18265},
      archivePrefix={arXiv},
      primaryClass={cs.CV},
      url={https://arxiv.org/abs/2508.18265}, 
}

@misc{Bai2025Qwen25VLTR,
      title={Qwen2.5-VL Technical Report}, 
      author={Shuai Bai and Keqin Chen and Xuejing Liu and Jialin Wang and Wenbin Ge and Sibo Song and Kai Dang and Peng Wang and Shijie Wang and Jun Tang and Humen Zhong and Yuanzhi Zhu and Mingkun Yang and Zhaohai Li and Jianqiang Wan and Pengfei Wang and Wei Ding and Zheren Fu and Yiheng Xu and Jiabo Ye and Xi Zhang and Tianbao Xie and Zesen Cheng and Hang Zhang and Zhibo Yang and Haiyang Xu and Junyang Lin},
      year={2025},
      eprint={2502.13923},
      archivePrefix={arXiv},
      primaryClass={cs.CV},
      url={https://arxiv.org/abs/2502.13923}, 
}

@misc{Yao2024MiniCPMVAG,
      title={MiniCPM-V: A GPT-4V Level MLLM on Your Phone}, 
      author={Yuan Yao and Tianyu Yu and Ao Zhang and Chongyi Wang and Junbo Cui and Hongji Zhu and Tianchi Cai and Haoyu Li and Weilin Zhao and Zhihui He and Qianyu Chen and Huarong Zhou and Zhensheng Zou and Haoye Zhang and Shengding Hu and Zhi Zheng and Jie Zhou and Jie Cai and Xu Han and Guoyang Zeng and Dahai Li and Zhiyuan Liu and Maosong Sun},
      year={2024},
      eprint={2408.01800},
      archivePrefix={arXiv},
      primaryClass={cs.CV},
      url={https://arxiv.org/abs/2408.01800}, 
}

@inproceedings{Zhang2025BeeAH,
 author = {Zhang, Yi and Ni, Bolin and Chen, Xin-Sheng and Zhang, Hengrui and Rao, Yongming and Peng, Houwen and Lu, Qinglin and Hu, Winston and Guo, Meng-Hao and Hu, Shi-min},
 booktitle = {International Conference on Learning Representations},
 editor = {C. Vondrick and B. Hariharan and C. Raffel and L. Pinto and D. Yang and A. Faust},
 pages = {13769--13836},
 title = {Bee: A High-Quality Corpus and Full-Stack Suite to Unlock Advanced Fully Open MLLMs},
 url = {https://proceedings.iclr.cc/paper_files/paper/2026/file/171291d8fed723c6dfc76330aa827ff8-Paper-Conference.pdf},
 volume = {2026},
 year = {2026}
}

@article{Li2024LLaVAOneVisionEV,
title = "LLaVA-OneVision: Easy Visual Task Transfer",
author = "Bo Li and Yuanhan Zhang and Dong Guo and Renrui Zhang and Feng Li and Hao Zhang and Kaichen Zhang and Peiyuan Zhang and Yanwei Li and Ziwei Liu and Chunyuan Li",
note = "Publisher Copyright: {\textcopyright} 2025, Transactions on Machine Learning Research. All rights reserved.",
year = "2025",
language = "English",
volume = "2025",
journal = "Transactions on Machine Learning Research",
issn = "2835-8856",
publisher = "Transactions on Machine Learning Research",
}

@misc{Kamath2025Gemma3T,
      title={Gemma 3 Technical Report}, 
      author={Gemma Team and Aishwarya Kamath and Johan Ferret and Shreya Pathak and Nino Vieillard and Ramona Merhej and Sarah Perrin and Tatiana Matejovicova and Alexandre Ramé and Morgane Rivière and Louis Rouillard and Thomas Mesnard and Geoffrey Cideron and Jean-bastien Grill and Sabela Ramos and Edouard Yvinec and Michelle Casbon and Etienne Pot and Ivo Penchev and Gaël Liu and Francesco Visin and Kathleen Kenealy and Lucas Beyer and Xiaohai Zhai and Anton Tsitsulin and Robert Busa-Fekete and Alex Feng and Noveen Sachdeva and Benjamin Coleman and Yi Gao and Basil Mustafa and Iain Barr and Emilio Parisotto and David Tian and Matan Eyal and Colin Cherry and Jan-Thorsten Peter and Danila Sinopalnikov and Surya Bhupatiraju and Rishabh Agarwal and Mehran Kazemi and Dan Malkin and Ravin Kumar and David Vilar and Idan Brusilovsky and Jiaming Luo and Andreas Steiner and Abe Friesen and Abhanshu Sharma and Abheesht Sharma and Adi Mayrav Gilady and Adrian Goedeckemeyer and Alaa Saade and Alex Feng and Alexander Kolesnikov and Alexei Bendebury and Alvin Abdagic and Amit Vadi and András György and André Susano Pinto and Anil Das and Ankur Bapna and Antoine Miech and Antoine Yang and Antonia Paterson and Ashish Shenoy and Ayan Chakrabarti and Bilal Piot and Bo Wu and Bobak Shahriari and Bryce Petrini and Charlie Chen and Charline Le Lan and Christopher A. Choquette-Choo and CJ Carey and Cormac Brick and Daniel Deutsch and Danielle Eisenbud and Dee Cattle and Derek Cheng and Dimitris Paparas and Divyashree Shivakumar Sreepathihalli and Doug Reid and Dustin Tran and Dustin Zelle and Eric Noland and Erwin Huizenga and Eugene Kharitonov and Frederick Liu and Gagik Amirkhanyan and Glenn Cameron and Hadi Hashemi and Hanna Klimczak-Plucińska and Harman Singh and Harsh Mehta and Harshal Tushar Lehri and Hussein Hazimeh and Ian Ballantyne and Idan Szpektor and Ivan Nardini and Jean Pouget-Abadie and Jetha Chan and Joe Stanton and John Wieting and Jonathan Lai and Jordi Orbay and Joseph Fernandez and Josh Newlan and Ju-yeong Ji and Jyotinder Singh and Kat Black and Kathy Yu and Kevin Hui and Kiran Vodrahalli and Klaus Greff and Linhai Qiu and Marcella Valentine and Marina Coelho and Marvin Ritter and Matt Hoffman and Matthew Watson and Mayank Chaturvedi and Michael Moynihan and Min Ma and Nabila Babar and Natasha Noy and Nathan Byrd and Nick Roy and Nikola Momchev and Nilay Chauhan and Noveen Sachdeva and Oskar Bunyan and Pankil Botarda and Paul Caron and Paul Kishan Rubenstein and Phil Culliton and Philipp Schmid and Pier Giuseppe Sessa and Pingmei Xu and Piotr Stanczyk and Pouya Tafti and Rakesh Shivanna and Renjie Wu and Renke Pan and Reza Rokni and Rob Willoughby and Rohith Vallu and Ryan Mullins and Sammy Jerome and Sara Smoot and Sertan Girgin and Shariq Iqbal and Shashir Reddy and Shruti Sheth and Siim Põder and Sijal Bhatnagar and Sindhu Raghuram Panyam and Sivan Eiger and Susan Zhang and Tianqi Liu and Trevor Yacovone and Tyler Liechty and Uday Kalra and Utku Evci and Vedant Misra and Vincent Roseberry and Vlad Feinberg and Vlad Kolesnikov and Woohyun Han and Woosuk Kwon and Xi Chen and Yinlam Chow and Yuvein Zhu and Zichuan Wei and Zoltan Egyed and Victor Cotruta and Minh Giang and Phoebe Kirk and Anand Rao and Kat Black and Nabila Babar and Jessica Lo and Erica Moreira and Luiz Gustavo Martins and Omar Sanseviero and Lucas Gonzalez and Zach Gleicher and Tris Warkentin and Vahab Mirrokni and Evan Senter and Eli Collins and Joelle Barral and Zoubin Ghahramani and Raia Hadsell and Yossi Matias and D. Sculley and Slav Petrov and Noah Fiedel and Noam Shazeer and Oriol Vinyals and Jeff Dean and Demis Hassabis and Koray Kavukcuoglu and Clement Farabet and Elena Buchatskaya and Jean-Baptiste Alayrac and Rohan Anil and Dmitry and Lepikhin and Sebastian Borgeaud and Olivier Bachem and Armand Joulin and Alek Andreev and Cassidy Hardin and Robert Dadashi and Léonard Hussenot},
      year={2025},
      eprint={2503.19786},
      archivePrefix={arXiv},
      primaryClass={cs.CL},
      url={https://arxiv.org/abs/2503.19786}, 
}

@article{Agrawal2024Pixtral1,
  author       = {Pravesh Agrawal and
                  Szymon Antoniak and
                  Emma Bou Hanna and
                  Baptiste Bout and
                  Devendra Singh Chaplot and
                  Jessica Chudnovsky and
                  Diogo Costa and
                  Baudouin De Monicault and
                  Saurabh Garg and
                  Th{\'{e}}ophile Gervet and
                  Soham Ghosh and
                  Am{\'{e}}lie H{\'{e}}liou and
                  Paul Jacob and
                  Albert Q. Jiang and
                  Kartik Khandelwal and
                  Timoth{\'{e}}e Lacroix and
                  Guillaume Lample and
                  Diego de Las Casas and
                  Thibaut Lavril and
                  Teven Le Scao and
                  Andy Lo and
                  William Marshall and
                  Louis Martin and
                  Arthur Mensch and
                  Pavankumar Muddireddy and
                  Valera Nemychnikova and
                  Marie Pellat and
                  Patrick von Platen and
                  Nikhil Raghuraman and
                  Baptiste Rozi{\`{e}}re and
                  Alexandre Sablayrolles and
                  Lucile Saulnier and
                  Romain Sauvestre and
                  Wendy Shang and
                  Roman Soletskyi and
                  Lawrence Stewart and
                  Pierre Stock and
                  Joachim Studnia and
                  Sandeep Subramanian and
                  Sagar Vaze and
                  Thomas Wang and
                  Sophia Yang},
  title        = {Pixtral 12B},
  journal      = {CoRR},
  volume       = {abs/2410.07073},
  year         = {2024},
  url          = {https://doi.org/10.48550/arXiv.2410.07073},
  doi          = {10.48550/ARXIV.2410.07073},
  eprinttype   = {arXiv},
  eprint       = {2410.07073},
  bibsource    = {dblp computer science bibliography, https://dblp.org}
}

@misc{Abouelenin2025Phi4MiniTR,
      title={Phi-4-Mini Technical Report: Compact yet Powerful Multimodal Language Models via Mixture-of-LoRAs}, 
      author={Abdelrahman Abouelenin and Atabak Ashfaq and Adam Atkinson and Hany Awadalla and Nguyen Bach and Jianmin Bao and Alon Benhaim and Martin Cai and Vishrav Chaudhary and Congcong Chen and Dong Chen and Dongdong Chen and Junkun Chen and Weizhu Chen and Yen-Chun Chen and Yi-ling Chen and Qi Dai and Xiyang Dai and Ruchao Fan and Mei Gao and Min Gao and Amit Garg and Abhishek Goswami and Junheng Hao and Amr Hendy and Yuxuan Hu and Xin Jin and Mahmoud Khademi and Dongwoo Kim and Young Jin Kim and Gina Lee and Jinyu Li and Yunsheng Li and Chen Liang and Xihui Lin and Zeqi Lin and Mengchen Liu and Yang Liu and Gilsinia Lopez and Chong Luo and Piyush Madan and Vadim Mazalov and Arindam Mitra and Ali Mousavi and Anh Nguyen and Jing Pan and Daniel Perez-Becker and Jacob Platin and Thomas Portet and Kai Qiu and Bo Ren and Liliang Ren and Sambuddha Roy and Ning Shang and Yelong Shen and Saksham Singhal and Subhojit Som and Xia Song and Tetyana Sych and Praneetha Vaddamanu and Shuohang Wang and Yiming Wang and Zhenghao Wang and Haibin Wu and Haoran Xu and Weijian Xu and Yifan Yang and Ziyi Yang and Donghan Yu and Ishmam Zabir and Jianwen Zhang and Li Lyna Zhang and Yunan Zhang and Xiren Zhou},
      year={2025},
      eprint={2503.01743},
      archivePrefix={arXiv},
      primaryClass={cs.CL},
      url={https://arxiv.org/abs/2503.01743}, 
}

@misc{Laurenon2024BuildingAB,
      title={Building and better understanding vision-language models: insights and future directions}, 
      author={Hugo Laurençon and Andrés Marafioti and Victor Sanh and Léo Tronchon},
      year={2024},
      eprint={2408.12637},
      archivePrefix={arXiv},
      primaryClass={cs.CV},
      url={https://arxiv.org/abs/2408.12637}, 
}

@misc{Abdin2024Phi3TR,
      title={Phi-3 Technical Report: A Highly Capable Language Model Locally on Your Phone}, 
      author={Marah Abdin and Jyoti Aneja and Hany Awadalla and Ahmed Awadallah and Ammar Ahmad Awan and Nguyen Bach and Amit Bahree and Arash Bakhtiari and Jianmin Bao and Harkirat Behl and Alon Benhaim and Misha Bilenko and Johan Bjorck and Sébastien Bubeck and Martin Cai and Qin Cai and Vishrav Chaudhary and Dong Chen and Dongdong Chen and Weizhu Chen and Yen-Chun Chen and Yi-Ling Chen and Hao Cheng and Parul Chopra and Xiyang Dai and Matthew Dixon and Ronen Eldan and Victor Fragoso and Jianfeng Gao and Mei Gao and Min Gao and Amit Garg and Allie Del Giorno and Abhishek Goswami and Suriya Gunasekar and Emman Haider and Junheng Hao and Russell J. Hewett and Wenxiang Hu and Jamie Huynh and Dan Iter and Sam Ade Jacobs and Mojan Javaheripi and Xin Jin and Nikos Karampatziakis and Piero Kauffmann and Mahoud Khademi and Dongwoo Kim and Young Jin Kim and Lev Kurilenko and James R. Lee and Yin Tat Lee and Yuanzhi Li and Yunsheng Li and Chen Liang and Lars Liden and Xihui Lin and Zeqi Lin and Ce Liu and Liyuan Liu and Mengchen Liu and Weishung Liu and Xiaodong Liu and Chong Luo and Piyush Madan and Ali Mahmoudzadeh and David Majercak and Matt Mazzola and Caio César Teodoro Mendes and Arindam Mitra and Hardik Modi and Anh Nguyen and Brandon Norick and Barun Patra and Daniel Perez-Becker and Thomas Portet and Reid Pryzant and Heyang Qin and Marko Radmilac and Liliang Ren and Gustavo de Rosa and Corby Rosset and Sambudha Roy and Olatunji Ruwase and Olli Saarikivi and Amin Saied and Adil Salim and Michael Santacroce and Shital Shah and Ning Shang and Hiteshi Sharma and Yelong Shen and Swadheen Shukla and Xia Song and Masahiro Tanaka and Andrea Tupini and Praneetha Vaddamanu and Chunyu Wang and Guanhua Wang and Lijuan Wang and Shuohang Wang and Xin Wang and Yu Wang and Rachel Ward and Wen Wen and Philipp Witte and Haiping Wu and Xiaoxia Wu and Michael Wyatt and Bin Xiao and Can Xu and Jiahang Xu and Weijian Xu and Jilong Xue and Sonali Yadav and Fan Yang and Jianwei Yang and Yifan Yang and Ziyi Yang and Donghan Yu and Lu Yuan and Chenruidong Zhang and Cyril Zhang and Jianwen Zhang and Li Lyna Zhang and Yi Zhang and Yue Zhang and Yunan Zhang and Xiren Zhou},
      year={2024},
      eprint={2404.14219},
      archivePrefix={arXiv},
      primaryClass={cs.CL},
      url={https://arxiv.org/abs/2404.14219}, 
}

@InProceedings{liu2023grounding,
author="Liu, Shilong
and Zeng, Zhaoyang
and Ren, Tianhe
and Li, Feng
and Zhang, Hao
and Yang, Jie
and Jiang, Qing
and Li, Chunyuan
and Yang, Jianwei
and Su, Hang
and Zhu, Jun
and Zhang, Lei",
editor="Leonardis, Ale{\v{s}}
and Ricci, Elisa
and Roth, Stefan
and Russakovsky, Olga
and Sattler, Torsten
and Varol, G{\"u}l",
title="Grounding DINO: Marrying DINO with Grounded Pre-training for Open-Set Object Detection",
booktitle="Computer Vision -- ECCV 2024",
year="2024",
publisher="Springer Nature Switzerland",
address="Cham",
pages="38--55",
isbn="978-3-031-72970-6"
}

@INPROCEEDINGS{Varghese2024YOLOv8AN,
  author={Varghese, Rejin and M., Sambath},
  booktitle={2024 International Conference on Advances in Data Engineering and Intelligent Computing Systems (ADICS)}, 
  title={YOLOv8: A Novel Object Detection Algorithm with Enhanced Performance and Robustness}, 
  year={2024},
  volume={},
  number={},
  pages={1-6},
  doi={10.1109/ADICS58448.2024.10533619}}

@InProceedings{Radford2021LearningTV,
  title = 	 {Learning Transferable Visual Models From Natural Language Supervision},
  author =       {Radford, Alec and Kim, Jong Wook and Hallacy, Chris and Ramesh, Aditya and Goh, Gabriel and Agarwal, Sandhini and Sastry, Girish and Askell, Amanda and Mishkin, Pamela and Clark, Jack and Krueger, Gretchen and Sutskever, Ilya},
  booktitle = 	 {Proceedings of the 38th International Conference on Machine Learning},
  pages = 	 {8748--8763},
  year = 	 {2021},
  editor = 	 {Meila, Marina and Zhang, Tong},
  volume = 	 {139},
  series = 	 {Proceedings of Machine Learning Research},
  month = 	 {18--24 Jul},
  publisher =    {PMLR},
  url = 	 {https://proceedings.mlr.press/v139/radford21a.html}
}

@inproceedings{Chen2024M3EmbeddingMM,
    title = "{M}3-Embedding: Multi-Linguality, Multi-Functionality, Multi-Granularity Text Embeddings Through Self-Knowledge Distillation",
    author = "Chen, Jianlyu  and
      Xiao, Shitao  and
      Zhang, Peitian  and
      Luo, Kun  and
      Lian, Defu  and
      Liu, Zheng",
    editor = "Ku, Lun-Wei  and
      Martins, Andre  and
      Srikumar, Vivek",
    booktitle = "Findings of the Association for Computational Linguistics: ACL 2024",
    month = aug,
    year = "2024",
    address = "Bangkok, Thailand",
    publisher = "Association for Computational Linguistics",
    url = "https://aclanthology.org/2024.findings-acl.137/",
    doi = "10.18653/v1/2024.findings-acl.137",
    pages = "2318--2335"
}

@InProceedings{Liu2023ImprovedBW,
    author    = {Liu, Haotian and Li, Chunyuan and Li, Yuheng and Lee, Yong Jae},
    title     = {Improved Baselines with Visual Instruction Tuning},
    booktitle = {Proceedings of the IEEE/CVF Conference on Computer Vision and Pattern Recognition (CVPR)},
    month     = {June},
    year      = {2024},
    pages     = {26296-26306}
}

@inproceedings{Zhu2023MiniGPT4EV,
  author       = {Deyao Zhu and
                  Jun Chen and
                  Xiaoqian Shen and
                  Xiang Li and
                  Mohamed Elhoseiny},
  title        = {MiniGPT-4: Enhancing Vision-Language Understanding with Advanced Large
                  Language Models},
  booktitle    = {The Twelfth International Conference on Learning Representations,
                  {ICLR} 2024, Vienna, Austria, May 7-11, 2024},
  publisher    = {OpenReview.net},
  year         = {2024},
  url          = {https://openreview.net/forum?id=1tZbq88f27},
  bibsource    = {dblp computer science bibliography, https://dblp.org}
}

@inproceedings{Zhang2023VideoLLaMAAI,
    title = "Video-{LL}a{MA}: An Instruction-tuned Audio-Visual Language Model for Video Understanding",
    author = "Zhang, Hang  and
      Li, Xin  and
      Bing, Lidong",
    editor = "Feng, Yansong  and
      Lefever, Els",
    booktitle = "Proceedings of the 2023 Conference on Empirical Methods in Natural Language Processing: System Demonstrations",
    month = dec,
    year = "2023",
    address = "Singapore",
    publisher = "Association for Computational Linguistics",
    url = "https://aclanthology.org/2023.emnlp-demo.49/",
    doi = "10.18653/v1/2023.emnlp-demo.49",
    pages = "543--553"
}

@InProceedings{Li2023MVBenchAC,
    author    = {Li, Kunchang and Wang, Yali and He, Yinan and Li, Yizhuo and Wang, Yi and Liu, Yi and Wang, Zun and Xu, Jilan and Chen, Guo and Luo, Ping and Wang, Limin and Qiao, Yu},
    title     = {MVBench: A Comprehensive Multi-modal Video Understanding Benchmark},
    booktitle = {Proceedings of the IEEE/CVF Conference on Computer Vision and Pattern Recognition (CVPR)},
    month     = {June},
    year      = {2024},
    pages     = {22195-22206}
}

@InProceedings{Fu2024VideoMMETF,
    author    = {Fu, Chaoyou and Dai, Yuhan and Luo, Yongdong and Li, Lei and Ren, Shuhuai and Zhang, Renrui and Wang, Zihan and Zhou, Chenyu and Shen, Yunhang and Zhang, Mengdan and Chen, Peixian and Li, Yanwei and Lin, Shaohui and Zhao, Sirui and Li, Ke and Xu, Tong and Zheng, Xiawu and Chen, Enhong and Shan, Caifeng and He, Ran and Sun, Xing},
    title     = {Video-MME: The First-Ever Comprehensive Evaluation Benchmark of Multi-modal LLMs in Video Analysis},
    booktitle = {Proceedings of the IEEE/CVF Conference on Computer Vision and Pattern Recognition (CVPR)},
    month     = {June},
    year      = {2025},
    pages     = {24108-24118}
}

@InProceedings{Zhou2024MLVUBM,
    author    = {Zhou, Junjie and Shu, Yan and Zhao, Bo and Wu, Boya and Liang, Zhengyang and Xiao, Shitao and Qin, Minghao and Yang, Xi and Xiong, Yongping and Zhang, Bo and Huang, Tiejun and Liu, Zheng},
    title     = {MLVU: Benchmarking Multi-task Long Video Understanding},
    booktitle = {Proceedings of the IEEE/CVF Conference on Computer Vision and Pattern Recognition (CVPR)},
    month     = {June},
    year      = {2025},
    pages     = {13691-13701}
}

@inproceedings{Wu2024LongVideoBenchAB,
 author = {Wu, Haoning and Li, Dongxu and Chen, Bei and Li, Junnan},
 booktitle = {Advances in Neural Information Processing Systems},
 doi = {10.52202/079017-0907},
 editor = {A. Globerson and L. Mackey and D. Belgrave and A. Fan and U. Paquet and J. Tomczak and C. Zhang},
 pages = {28828--28857},
 publisher = {Curran Associates, Inc.},
 title = {LongVideoBench: A Benchmark for Long-context Interleaved Video-Language Understanding},
 url = {https://proceedings.neurips.cc/paper_files/paper/2024/file/329ad516cf7a6ac306f29882e9c77558-Paper-Datasets_and_Benchmarks_Track.pdf},
 volume = {37},
 year = {2024}
}

@InProceedings{Wang2024VideoAgentLV,
author="Wang, Xiaohan
and Zhang, Yuhui
and Zohar, Orr
and Yeung-Levy, Serena",
editor="Leonardis, Ale{\v{s}}
and Ricci, Elisa
and Roth, Stefan
and Russakovsky, Olga
and Sattler, Torsten
and Varol, G{\"u}l",
title="VideoAgent: Long-Form Video Understanding with Large Language Model as Agent",
booktitle="Computer Vision -- ECCV 2024",
year="2025",
publisher="Springer Nature Switzerland",
address="Cham",
pages="58--76",
isbn="978-3-031-72989-8"
}

@InProceedings{Suris2023ViperGPTVI,
    author    = {Sur{\'\i}s, D{\'\i}dac and Menon, Sachit and Vondrick, Carl},
    title     = {ViperGPT: Visual Inference via Python Execution for Reasoning},
    booktitle = {Proceedings of the IEEE/CVF International Conference on Computer Vision (ICCV)},
    month     = {October},
    year      = {2023},
    pages     = {11888-11898}
}

@article{Li2023VideoChatCV,
  author       = {Kunchang Li and
                  Yinan He and
                  Yi Wang and
                  Yizhuo Li and
                  Wenhai Wang and
                  Ping Luo and
                  Yali Wang and
                  Limin Wang and
                  Yu Qiao},
  title        = {VideoChat: chat-centric video understanding},
  journal      = {Sci. China Inf. Sci.},
  volume       = {68},
  number       = {10},
  year         = {2025},
  url          = {https://doi.org/10.1007/s11432-024-4321-9},
  doi          = {10.1007/S11432-024-4321-9},
  bibsource    = {dblp computer science bibliography, https://dblp.org}
}

@inproceedings{Maaz2023VideoChatGPTTD,
    title = "Video-{C}hat{GPT}: Towards Detailed Video Understanding via Large Vision and Language Models",
    author = "Maaz, Muhammad  and
      Rasheed, Hanoona  and
      Khan, Salman  and
      Khan, Fahad",
    editor = "Ku, Lun-Wei  and
      Martins, Andre  and
      Srikumar, Vivek",
    booktitle = "Proceedings of the 62nd Annual Meeting of the Association for Computational Linguistics (Volume 1: Long Papers)",
    month = aug,
    year = "2024",
    address = "Bangkok, Thailand",
    publisher = "Association for Computational Linguistics",
    url = "https://aclanthology.org/2024.acl-long.679/",
    doi = "10.18653/v1/2024.acl-long.679",
    pages = "12585--12602"
}

@InProceedings{Yang2024VCAVC,
    author    = {Yang, Zeyuan and Chen, Delin and Yu, Xueyang and Shen, Maohao and Gan, Chuang},
    title     = {VCA: Video Curious Agent for Long Video Understanding},
    booktitle = {Proceedings of the IEEE/CVF International Conference on Computer Vision (ICCV)},
    month     = {October},
    year      = {2025},
    pages     = {20168-20179}
}

@article{Zhi2025VideoAgent2ET,
  author       = {Zhuo Zhi and
                  Qiangqiang Wu and
                  Minghe shen and
                  Wenbo Li and
                  Yinchuan Li and
                  Kun Shao and
                  Kaiwen Zhou},
  title        = {VideoAgent2: Enhancing the LLM-Based Agent System for Long-Form Video
                  Understanding by Uncertainty-Aware CoT},
  journal      = {CoRR},
  volume       = {abs/2504.04471},
  year         = {2025},
  url          = {https://doi.org/10.48550/arXiv.2504.04471},
  doi          = {10.48550/ARXIV.2504.04471},
  eprinttype   = {arXiv},
  eprint       = {2504.04471},
  bibsource    = {dblp computer science bibliography, https://dblp.org}
}

@misc{Li2026SkillsBenchBH,
      title={SkillsBench: Benchmarking How Well Agent Skills Work Across Diverse Tasks}, 
      author={Xiangyi Li and Yimin Liu and Wenbo Chen and Bingran You and Zonglin Di and Yifeng He and Shenghan Zheng and Kyoung Whan Choe and Jiankai Sun and Shuyi Wang and Chujun Tao and Binxu Li and Xuandong Zhao and Hejia Geng and Xiaojun Wu and Junwei Zhou and Xiaokun Chen and Hanwen Xing and Yubo Li and Qunhong Zeng and Di Wang and Yuanli Wang and Roey Ben Chaim and Penghao Jiang and Haotian Shen and Luyang Kong and Xinyi Liu and Runhui Wang and Xuanqing Liu and Jiachen Li and Xin Lan and Yueqian Lin and Wengao Ye and Junwei He and Songlin Li and Yue Zhang and Yipeng Gao and Yijiang Li and Ze Ma and Liqiang Jing and Tianyu Wang and Kaixin Li and Yiqi Xue and Haoran Lyu and Yizhuo He and Yuchen Tian and Shutong Wu and Bowei Wang and Yixuan Gao and Bo Chen and Litong Liu and Sikai Cheng and Jiajun Bao and Shuaicheng Tong and Shuwen Xu and Terry Yue Zhuo and Tinghan Ye and Qi Qi and Miao Li and Longtai Liao and Zelin Tan and Chang Shi and Xilin Tang and Srinath Tankasala and Boqin Yuan and Yaoyao Qian and Jianhong Tu and Chenguang Wang and Yizhou Sun and Wei Wang and Aaron Taylor and Ziyue Yang and Changkun Guan and Zhikang Dong and Xinyu Zhang and Steven Dillmann and Han-chung Lee and Dawn Song},
      year={2026},
      eprint={2602.12670},
      archivePrefix={arXiv},
      primaryClass={cs.AI},
      url={https://arxiv.org/abs/2602.12670}, 
}

@misc{deepseekai2026deepseekv4,
      title={DeepSeek-V4: Towards Highly Efficient Million-Token Context Intelligence},
      author={DeepSeek-AI},
      year={2026},
}

@inproceedings{DBLP:conf/iclr/BurgessWZRLDDY25,
  author       = {James Burgess and
                  Xiaohan Wang and
                  Yuhui Zhang and
                  Anita Rau and
                  Alejandro Lozano and
                  Lisa Dunlap and
                  Trevor Darrell and
                  Serena Yeung{-}Levy},
  title        = {Video Action Differencing},
  booktitle    = {The Thirteenth International Conference on Learning Representations,
                  {ICLR} 2025, Singapore, April 24-28, 2025},
  publisher    = {OpenReview.net},
  year         = {2025},
  url          = {https://openreview.net/forum?id=3bcN6xlO6f},
  bibsource    = {dblp computer science bibliography, https://dblp.org}
}

@InProceedings{Nagarajan_2024_CVPR,
    author    = {Nagarajan, Tushar and Torresani, Lorenzo},
    title     = {Step Differences in Instructional Video},
    booktitle = {Proceedings of the IEEE/CVF Conference on Computer Vision and Pattern Recognition (CVPR)},
    month     = {June},
    year      = {2024},
    pages     = {18740-18750}
}

@inproceedings{DBLP:conf/iccv/QiuSYEMFHS25,
  author       = {Yue Qiu and
                  Yanjun Sun and
                  Takuma Yagi and
                  Shusaku Egami and
                  Natsuki Miyata and
                  Ken Fukuda and
                  Kensho Hara and
                  Ryusuke Sagawa},
  title        = {VideoSetDiff: Identifying and Reasoning Similarities and Differences
                  in Similar Videos},
  booktitle    = {{IEEE/CVF} International Conference on Computer Vision, {ICCV} 2025,
                  Honolulu, HI, USA, October 19-25, 2025},
  pages        = {12242--12252},
  publisher    = {{IEEE}},
  year         = {2025},
  url          = {https://doi.org/10.1109/ICCV51701.2025.01138},
  doi          = {10.1109/ICCV51701.2025.01138},
  bibsource    = {dblp computer science bibliography, https://dblp.org}
}

@inproceedings{DBLP:conf/nips/LiberatoriCVWRR25,
  author       = {Benedetta Liberatori and
                  Alessandro Conti and
                  Lorenzo Vaquero and
                  Yiming Wang and
                  Elisa Ricci and
                  Paolo Rota},
  editor       = {Danielle Belgrave and
                  Cheng Zhang and
                  Laura N. Montoya and
                  Hsuan{-}Tien Lin and
                  Razvan Pascanu and
                  Piotr Koniusz and
                  Marzyeh Ghassemi and
                  Nancy Chen and
                  Iv{\'{a}}n Vladimir Meza Ru{\'{\i}}z and
                  Arturo Loaiza{-}Bonilla},
  title        = {ConViS-Bench: Estimating Video Similarity Through Semantic Concepts},
  booktitle    = {Advances in Neural Information Processing Systems 38: Annual Conference
                  on Neural Information Processing Systems 2025, NeurIPS 2025, San Diego,
                  CA, USA, December 2-7, 2025 / Mexico City, Mexico, November 30 - December
                  5, 2025},
  year         = {2025},
  url          = {http://papers.nips.cc/paper\_files/paper/2025/hash/e33a78b8e849e48f824be5ef8bee87e7-Abstract-Datasets\_and\_Benchmarks\_Track.html},
  bibsource    = {dblp computer science bibliography, https://dblp.org}
}

@inproceedings{ICLR2024_2862ccf0,
 author = {Zhu, Bin and Lin, Bin and Ning, Munan and YAN, YANG and Cui, Jiaxi and HongFa, WANG and Pang, Yatian and Jiang, Wenhao and Zhang, Junwu and Li, Zongwei and Zhang, Wancai and Li, Zhifeng and Liu, Wei and Li, Yuan},
 booktitle = {International Conference on Learning Representations},
 editor = {B. Kim and Y. Yue and S. Chaudhuri and K. Fragkiadaki and M. Khan and Y. Sun},
 pages = {9588--9608},
 title = {LanguageBind: Extending Video-Language Pretraining to N-modality by Language-based Semantic Alignment},
 url = {https://proceedings.iclr.cc/paper_files/paper/2024/file/2862ccf01e3843c81623b246895bcc45-Paper-Conference.pdf},
 volume = {2024},
 year = {2024}
}

@misc{singh2026openaigpt5card,
      title={OpenAI GPT-5 System Card}, 
      author={Aaditya Singh and Adam Fry and Adam Perelman and Adam Tart and Adi Ganesh and Ahmed El-Kishky and Aidan McLaughlin and Aiden Low and AJ Ostrow and Akhila Ananthram and Akshay Nathan and Alan Luo and Alec Helyar and Aleksander Madry and Aleksandr Efremov and Aleksandra Spyra and Alex Baker-Whitcomb and Alex Beutel and Alex Karpenko and Alex Makelov and Alex Neitz and Alex Wei and Alexandra Barr and Alexandre Kirchmeyer and Alexey Ivanov and Alexi Christakis and Alistair Gillespie and Allison Tam and Ally Bennett and Alvin Wan and Alyssa Huang and Amy McDonald Sandjideh and Amy Yang and Ananya Kumar and Andre Saraiva and Andrea Vallone and Andrei Gheorghe and Andres Garcia Garcia and Andrew Braunstein and Andrew Liu and Andrew Schmidt and Andrey Mereskin and Andrey Mishchenko and Andy Applebaum and Andy Rogerson and Ann Rajan and Annie Wei and Anoop Kotha and Anubha Srivastava and Anushree Agrawal and Arun Vijayvergiya and Ashley Tyra and Ashvin Nair and Avi Nayak and Ben Eggers and Bessie Ji and Beth Hoover and Bill Chen and Blair Chen and Boaz Barak and Borys Minaiev and Botao Hao and Bowen Baker and Brad Lightcap and Brandon McKinzie and Brandon Wang and Brendan Quinn and Brian Fioca and Brian Hsu and Brian Yang and Brian Yu and Brian Zhang and Brittany Brenner and Callie Riggins Zetino and Cameron Raymond and Camillo Lugaresi and Carolina Paz and Cary Hudson and Cedric Whitney and Chak Li and Charles Chen and Charlotte Cole and Chelsea Voss and Chen Ding and Chen Shen and Chengdu Huang and Chris Colby and Chris Hallacy and Chris Koch and Chris Lu and Christina Kaplan and Christina Kim and CJ Minott-Henriques and Cliff Frey and Cody Yu and Coley Czarnecki and Colin Reid and Colin Wei and Cory Decareaux and Cristina Scheau and Cyril Zhang and Cyrus Forbes and Da Tang and Dakota Goldberg and Dan Roberts and Dana Palmie and Daniel Kappler and Daniel Levine and Daniel Wright and Dave Leo and David Lin and David Robinson and Declan Grabb and Derek Chen and Derek Lim and Derek Salama and Dibya Bhattacharjee and Dimitris Tsipras and Dinghua Li and Dingli Yu and DJ Strouse and Drew Williams and Dylan Hunn and Ed Bayes and Edwin Arbus and Ekin Akyurek and Elaine Ya Le and Elana Widmann and Eli Yani and Elizabeth Proehl and Enis Sert and Enoch Cheung and Eri Schwartz and Eric Han and Eric Jiang and Eric Mitchell and Eric Sigler and Eric Wallace and Erik Ritter and Erin Kavanaugh and Evan Mays and Evgenii Nikishin and Fangyuan Li and Felipe Petroski Such and Filipe de Avila Belbute Peres and Filippo Raso and Florent Bekerman and Foivos Tsimpourlas and Fotis Chantzis and Francis Song and Francis Zhang and Gaby Raila and Garrett McGrath and Gary Briggs and Gary Yang and Giambattista Parascandolo and Gildas Chabot and Grace Kim and Grace Zhao and Gregory Valiant and Guillaume Leclerc and Hadi Salman and Hanson Wang and Hao Sheng and Haoming Jiang and Haoyu Wang and Haozhun Jin and Harshit Sikchi and Heather Schmidt and Henry Aspegren and Honglin Chen and Huida Qiu and Hunter Lightman and Ian Covert and Ian Kivlichan and Ian Silber and Ian Sohl and Ibrahim Hammoud and Ignasi Clavera and Ikai Lan and Ilge Akkaya and Ilya Kostrikov and Irina Kofman and Isak Etinger and Ishaan Singal and Jackie Hehir and Jacob Huh and Jacqueline Pan and Jake Wilczynski and Jakub Pachocki and James Lee and James Quinn and Jamie Kiros and Janvi Kalra and Jasmyn Samaroo and Jason Wang and Jason Wolfe and Jay Chen and Jay Wang and Jean Harb and Jeffrey Han and Jeffrey Wang and Jennifer Zhao and Jeremy Chen and Jerene Yang and Jerry Tworek and Jesse Chand and Jessica Landon and Jessica Liang and Ji Lin and Jiancheng Liu and Jianfeng Wang and Jie Tang and Jihan Yin and Joanne Jang and Joel Morris and Joey Flynn and Johannes Ferstad and Johannes Heidecke and John Fishbein and John Hallman and Jonah Grant and Jonathan Chien and Jonathan Gordon and Jongsoo Park and Jordan Liss and Jos Kraaijeveld and Joseph Guay and Joseph Mo and Josh Lawson and Josh McGrath and Joshua Vendrow and Joy Jiao and Julian Lee and Julie Steele and Julie Wang and Junhua Mao and Kai Chen and Kai Hayashi and Kai Xiao and Kamyar Salahi and Kan Wu and Karan Sekhri and Karan Sharma and Karan Singhal and Karen Li and Kenny Nguyen and Keren Gu-Lemberg and Kevin King and Kevin Liu and Kevin Stone and Kevin Yu and Kristen Ying and Kristian Georgiev and Kristie Lim and Kushal Tirumala and Kyle Miller and Lama Ahmad and Larry Lv and Laura Clare and Laurance Fauconnet and Lauren Itow and Lauren Yang and Laurentia Romaniuk and Leah Anise and Lee Byron and Leher Pathak and Leon Maksin and Leyan Lo and Leyton Ho and Li Jing and Liang Wu and Liang Xiong and Lien Mamitsuka and Lin Yang and Lindsay McCallum and Lindsey Held and Liz Bourgeois and Logan Engstrom and Lorenz Kuhn and Louis Feuvrier and Lu Zhang and Lucas Switzer and Lukas Kondraciuk and Lukasz Kaiser and Manas Joglekar and Mandeep Singh and Mandip Shah and Manuka Stratta and Marcus Williams and Mark Chen and Mark Sun and Marselus Cayton and Martin Li and Marvin Zhang and Marwan Aljubeh and Matt Nichols and Matthew Haines and Max Schwarzer and Mayank Gupta and Meghan Shah and Melody Y. Guan and Melody Huang and Meng Dong and Mengqing Wang and Mia Glaese and Micah Carroll and Michael Lampe and Michael Malek and Michael Sharman and Michael Zhang and Michele Wang and Michelle Pokrass and Mihai Florian and Mikhail Pavlov and Miles Wang and Ming Chen and Mingxuan Wang and Minnia Feng and Mo Bavarian and Molly Lin and Moose Abdool and Mostafa Rohaninejad and Nacho Soto and Natalie Staudacher and Natan LaFontaine and Nathan Marwell and Nelson Liu and Nick Preston and Nick Turley and Nicklas Ansman and Nicole Blades and Nikil Pancha and Nikita Mikhaylin and Niko Felix and Nikunj Handa and Nishant Rai and Nitish Keskar and Noam Brown and Ofir Nachum and Oleg Boiko and Oleg Murk and Olivia Watkins and Oona Gleeson and Pamela Mishkin and Patryk Lesiewicz and Paul Baltescu and Pavel Belov and Peter Zhokhov and Philip Pronin and Phillip Guo and Phoebe Thacker and Qi Liu and Qiming Yuan and Qinghua Liu and Rachel Dias and Rachel Puckett and Rahul Arora and Ravi Teja Mullapudi and Raz Gaon and Reah Miyara and Rennie Song and Rishabh Aggarwal and RJ Marsan and Robel Yemiru and Robert Xiong and Rohan Kshirsagar and Rohan Nuttall and Roman Tsiupa and Ronen Eldan and Rose Wang and Roshan James and Roy Ziv and Rui Shu and Ruslan Nigmatullin and Saachi Jain and Saam Talaie and Sam Altman and Sam Arnesen and Sam Toizer and Sam Toyer and Samuel Miserendino and Sandhini Agarwal and Sarah Yoo and Savannah Heon and Scott Ethersmith and Sean Grove and Sean Taylor and Sebastien Bubeck and Sever Banesiu and Shaokyi Amdo and Shengjia Zhao and Sherwin Wu and Shibani Santurkar and Shiyu Zhao and Shraman Ray Chaudhuri and Shreyas Krishnaswamy and Shuaiqi and Xia and Shuyang Cheng and Shyamal Anadkat and Simón Posada Fishman and Simon Tobin and Siyuan Fu and Somay Jain and Song Mei and Sonya Egoian and Spencer Kim and Spug Golden and SQ Mah and Steph Lin and Stephen Imm and Steve Sharpe and Steve Yadlowsky and Sulman Choudhry and Sungwon Eum and Suvansh Sanjeev and Tabarak Khan and Tal Stramer and Tao Wang and Tao Xin and Tarun Gogineni and Taya Christianson and Ted Sanders and Tejal Patwardhan and Thomas Degry and Thomas Shadwell and Tianfu Fu and Tianshi Gao and Timur Garipov and Tina Sriskandarajah and Toki Sherbakov and Tomek Korbak and Tomer Kaftan and Tomo Hiratsuka and Tongzhou Wang and Tony Song and Tony Zhao and Troy Peterson and Val Kharitonov and Victoria Chernova and Vineet Kosaraju and Vishal Kuo and Vitchyr Pong and Vivek Verma and Vlad Petrov and Wanning Jiang and Weixing Zhang and Wenda Zhou and Wenlei Xie and Wenting Zhan and Wes McCabe and Will DePue and Will Ellsworth and Wulfie Bain and Wyatt Thompson and Xiangning Chen and Xiangyu Qi and Xin Xiang and Xinwei Shi and Yann Dubois and Yaodong Yu and Yara Khakbaz and Yifan Wu and Yilei Qian and Yin Tat Lee and Yinbo Chen and Yizhen Zhang and Yizhong Xiong and Yonglong Tian and Young Cha and Yu Bai and Yu Yang and Yuan Yuan and Yuanzhi Li and Yufeng Zhang and Yuguang Yang and Yujia Jin and Yun Jiang and Yunyun Wang and Yushi Wang and Yutian Liu and Zach Stubenvoll and Zehao Dou and Zheng Wu and Zhigang Wang},
      year={2026},
      eprint={2601.03267},
      archivePrefix={arXiv},
      primaryClass={cs.CL},
      url={https://arxiv.org/abs/2601.03267}, 
}

@misc{qwen3.6-27b,
    title  = {{Qwen3.6-27B}: Flagship-Level Coding in a {27B} Dense Model},
    author = {{Qwen Team}},
    month  = {April},
    year   = {2026},
    url    = {https://qwen.ai/blog?id=qwen3.6-27b}
}

\appendix
\clearpage
\label{sec:appendix}

\section{Data Construction Details}
\label{appendix:data_construction_detailes}
MVX-Bench comprises 11 tasks, as shown in Figure~\ref{fig:data}, all of which have been repurposed into a multiple-choice question-answering format. These tasks utilize a diverse collection of videos from various sources. 

We use public datasets, pretrained models, and open-source tools only for research and evaluation purposes, following their original licenses, terms of use, and access conditions. For MVX-Bench, we do not redistribute raw videos unless permitted by the source licenses; otherwise, we release only derived annotations, metadata, or instructions for accessing the original sources. For models and tools used in SAMA, we cite their creators and follow their respective licenses. Our released artifacts will include license information for our annotations/code and references to the licenses of third-party resources. The MVX-Bench is publicly available at \url{https://huggingface.co/datasets/MVX-bench/MVX-Bench}.

% \begin{figure*}[ht!]
% \centering
% \includegraphics[width=\textwidth]{figures/MVA-data.pdf}
% \caption{Overview of the MVX-Bench.}
% \label{fig:data}
% \end{figure*}

\paragraph{Counting.}
This task evaluates whether LVLMs can accurately perceive object counts across multiple videos. 
Constructed from VideoCount~\cite{AminiNaieni2025OpenWorldOC}, which provides videos annotated with object categories and counts, we generate multiple-choice questions by specifying a target object and a required numerical relationship (same, greater, or less). 
Given a reference video, the model must identify which candidate video satisfies the specified relationship.
Questions are constructed through a template-based pipeline. 
Specifically, we first sample a reference video and use the annotated count of a target object. 
We then randomly select a numerical relationship (same, greater, or less) and sample four candidate videos with annotated object counts as answer options. 
The candidates are filtered to ensure that exactly one option satisfies the required numerical relationship with respect to the reference.
The multi-video setting requires precise quantitative perception for certain objects and cross-video comparison, exposing models’ weaknesses in object counting.

\paragraph{Forensic Detection.}
Recent advances in generative AI have raised concerns about malicious content creation, motivating the need for automatic detection of manipulated media. 
In this task, the model is presented with two videos and must determine which one has been manipulated.
The task is constructed based on WildDeepfake~\cite{Zi2020WildDeepfakeAC}, which provides videos labeled as either authentic or manipulated. 
For each question, we sample one real video and one manipulated video from the dataset and present them as candidate options for the MCQ.

\paragraph{Action Matching.} This task evaluates whether LVLMs can recognize and match actions across multiple videos. 
Given a reference video depicting a specific action, the model must identify which of the four candidate videos shows the same action.
The task is constructed based on Kinetics~\cite{Carreira2019ASN}, which provides videos annotated with action labels. 
For each question, we sample one reference video and select a positive candidate with the same action label as the correct answer. 
Three negative candidates are sampled from videos with different action labels.

\paragraph{Re-Identification.} This task evaluates whether LVLMs can determine if individuals appearing in different videos are the same person. 
Given two video clips captured from different cameras, the model must decide whether they depict the same individual.
The task is constructed based on MEVID~\cite{DBLP:conf/wacv/DavilaDLFPCCBMHC23}, which provides videos of unique individuals recorded under varying camera views, clothing, and backgrounds. 
For each question, we sample either a positive pair that shares the same identity or a negative pair consisting of different individuals.
Positive pairs share identity but differ in clothing, viewpoint, and scene context. 
This task requires disentangling identity from superficial appearance changes: the model must rely on cues such as body structure, gait, and facial characteristics while ignoring variations that could mislead naive matching strategies.

\paragraph{Art Style.}
This task evaluates LVLMs' ability to analyze and distinguish both local and global stylistic patterns across videos. 
Given a reference animation clip, the model must identify which of four candidate clips comes from the same show based solely on visual style.
The task is constructed based on Anim-400K~\cite{Cai2024Anim400KAL}, which provides animation clips from diverse shows. 
For each question, we sample a reference clip and select a positive candidate from the same show as the correct answer, while three negative candidates are sampled from different shows.
Unlike conventional content-matching tasks, this setting requires the model to ignore what is depicted (e.g., characters or scenes) and instead focus on how it is depicted, such as color palette, line quality, shading techniques, and animation fluidity.

\paragraph{Multi-View Reasoning.} This task evaluates whether LVLMs can reason about scene consistency across different viewpoints and sensing modalities. 
Given a reference video, the model must identify which candidate was captured at the same physical location or depicts the same event, despite variations in camera perspective or modality (e.g., RGB vs. depth).
The task is constructed based on the Free-Viewpoint Dataset~\cite{10403933}, which uses a 12-view synchronous camera system to record the same event from multiple viewpoints. 
For each question, we sample a reference video and select a positive candidate capturing the same event from a different viewpoint as the correct answer. 
Three negative candidates are sampled from different events.

\paragraph{Video Similarity.} This task evaluates whether LVLMs can perform fine-grained similarity judgments across videos, reflecting a nuanced understanding of visual features, patterns, and aesthetics. 
The task is constructed based on FIVR-200K~\cite{KordopatisZilos2018FIVRFI}, which provides human-annotated similarity relationships and graded similarity levels between videos. 
For each question, we select as the correct answer a candidate labeled as a near-duplicate of the reference video in the dataset. 
Three negative candidates are randomly sampled from videos that are not annotated as near-duplicates of the reference.
This setting requires distinguishing subtle variations in visual content while recognizing high-level structural and semantic similarity.

\paragraph{Counterfactual Reasoning.} This task evaluates whether LVLMs can reason about hypothetical scenarios that differ from observed events. 
It requires models to mentally simulate alternative outcomes based on visual and linguistic context.
Given a reference video paired with a counterfactual question and two answer options, the model must identify which candidate video is associated with the same correct answer option (e.g., Option 1 or Option 2) under its respective counterfactual question.
The task is constructed based on ACQUIRED~\cite{Wu2023ACQUIREDAD}, which provides videos annotated with counterfactual questions, two answer options, and ground-truth labels. 
For each question, we sample a reference video together with its counterfactual question and answer options. 
We then construct four candidate videos, each paired with its own counterfactual question and two answer options. 
The correct candidate is selected such that its ground-truth answer option (e.g., Option 1 or Option 2) matches that of the reference, while negative candidates are chosen to ensure their correct answer options differ from the reference.
The model is required to select the candidate whose correct answer option index matches that of the reference.

\paragraph{Defeasible Entailment.}
This task evaluates whether LVLMs can perform defeasible reasoning and dynamically update beliefs in light of new evidence. 
Given a textual hypothesis and a premise video, the model is further presented with an update video and must determine whether the new video evidence strengthens or weakens the entailment relationship between the premise and the hypothesis.
The task is constructed based on DVidE~\cite{Zhang2025CanVL}, which provides hypothesis texts, premise videos, update evidence, and corresponding labels indicating whether the update strengthens or weakens the original entailment. 
We use the hypothesis text and premise video as the initial context, replace the textual update with its corresponding update video, and require the model to predict whether the update video strengthens or weakens the entailment relation.
This setting evaluates dynamic belief revision: the model must first infer an initial relationship between the premise video and hypothesis, then assess how additional visual evidence modifies that judgment. It tests whether models can flexibly revise conclusions rather than anchoring to initial interpretations.

\paragraph{Sequence Reasoning.} This task evaluates whether LVLMs can perform temporal reasoning over event sequences. 
The model is given multiple video clips from the coherent activity in shuffled order and must reconstruct the correct chronological sequence.
The task is constructed based on CI-VID~\cite{Ju2025CIVIDAC}, which provides temporally ordered clips from the coherent activity. 
For each question, we sample several clips from a single event and randomly permute their order as input. 
The model is required to predict the correct temporal ordering.
Rather than relying on explicit timestamps or narration, the model must infer temporal structure from visual cues such as scene progression, action continuity, object state changes, and causal dependencies between events.

\paragraph{Spatial Relation.} 
This task evaluates whether LVLMs can understand and compare motion dynamics across videos. 
Given a reference video exhibiting specific motion trends (e.g., forward movement or lateral panning), the model must identify which candidate video shares at least one motion trend with the reference.
The task is constructed based on SpatialVID~\cite{Wang2025SpatialVIDAL}, which provides videos annotated with motion trend labels. 
For each question, we select a reference video and sample a positive candidate that shares at least one motion label with the reference. 
Three negative candidates are sampled from videos whose motion labels do not overlap with those of the reference. 
Positives are required to exhibit overlapping motion patterns, while negatives contain completely disjoint motion trends. 
This design requires the model to abstract away from visual content and focus on underlying spatial dynamics—how the camera or subjects move through space—when comparing videos.

\section{Case Analysis}
\begin{figure*}[ht!]
\centering
\includegraphics[width=\textwidth]{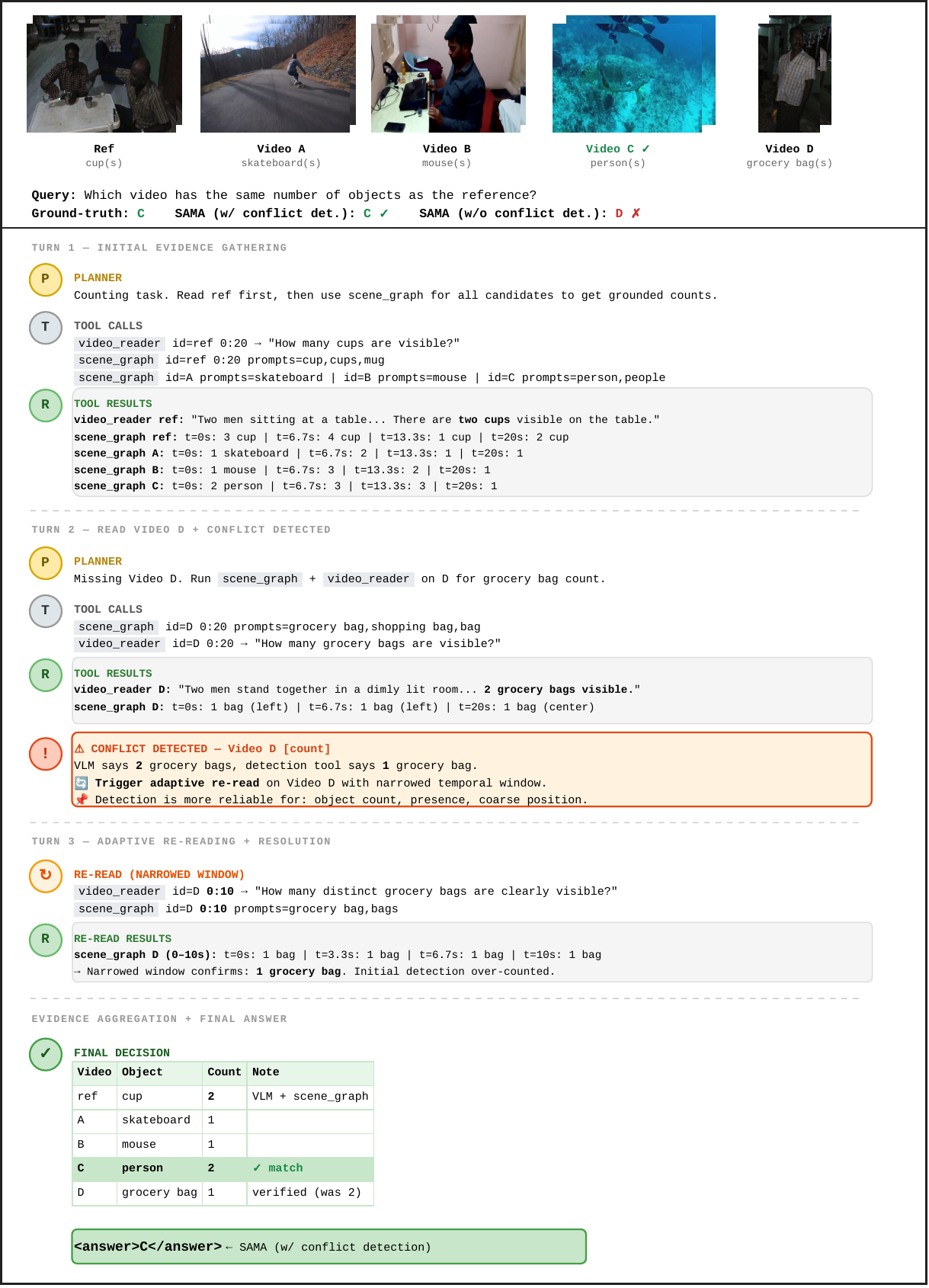}
\caption{Case study of the conflict detection mechanism.}
\label{fig:conflict_example}
\end{figure*}

Figure~\ref{fig:conflict_example} shows a representative example where the conflict detection mechanism directly changes the final prediction. The task requires identifying which video contains the same number of objects as the reference video (2 cups). When analyzing Video D, the VLM (\texttt{video\_reader}) reports two grocery bags, while the detection tool (\texttt{scene\_graph}) identifies one. This discrepancy automatically triggers the conflict detection module, which initiates an adaptive re-read with a narrowed temporal window (0--10s). The focused re-read corrects the count for Video D to one grocery bag, eliminating it as a candidate. With all counts verified, the agent correctly aligns the reference (2 cups) with Video C (2 persons). In contrast, without conflict detection, the initial overcount for Video D is never corrected, and the agent incorrectly selects Video D. This case illustrates that the verification layer not only improves evidence reliability but can also directly alter the final decision.

\begin{figure*}[ht!]
\centering
\includegraphics[width=\textwidth]{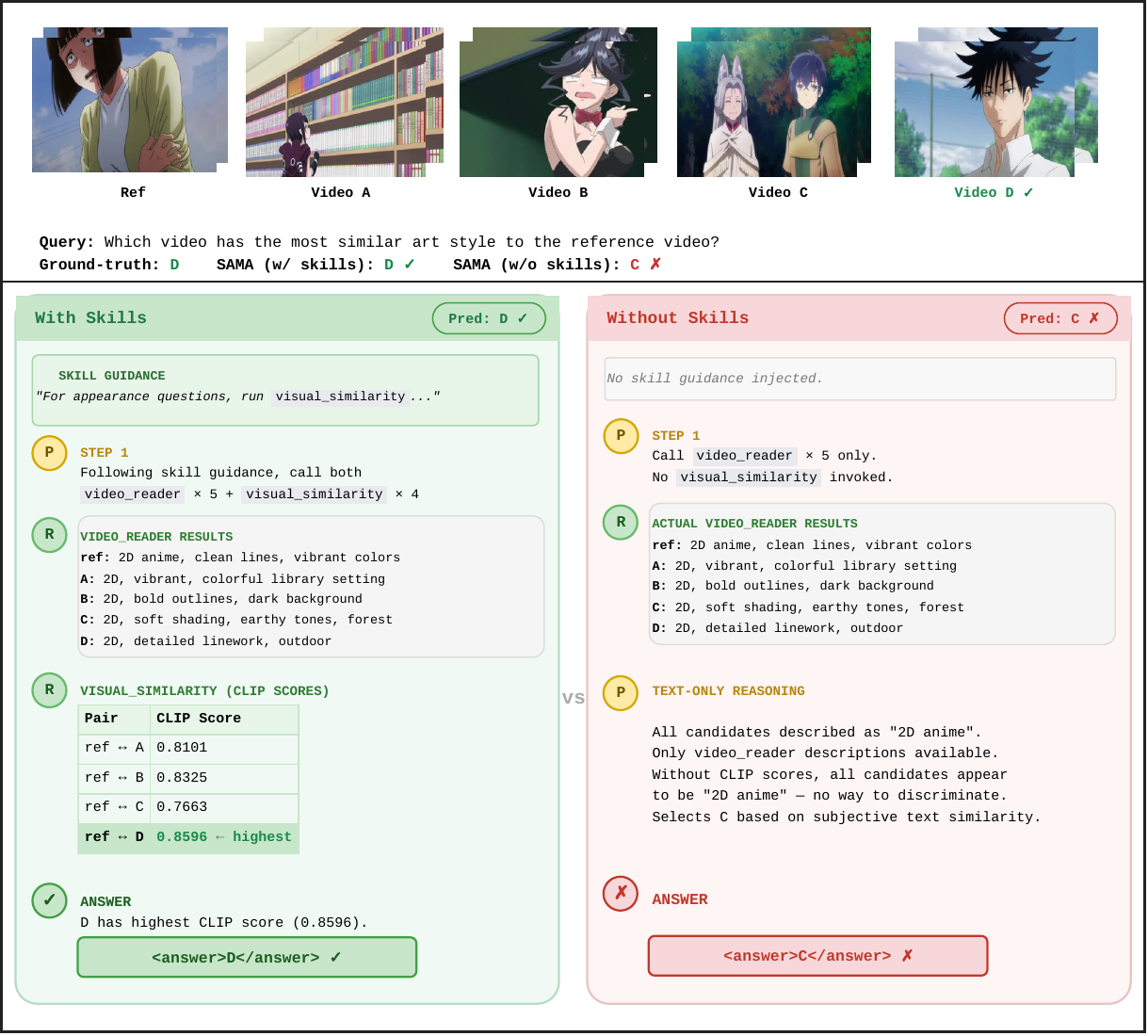}
\caption{Case study of skill-guided tool selection.}
\label{fig:skill_example}
\end{figure*}

Figure~\ref{fig:skill_example} presents a case where task-specific skills determine the correct tool strategy. The task asks which video shares the most similar art style with the reference. With skills enabled, the \textit{multi-video-compare} skill explicitly instructs the agent to invoke visual\_similarity for style-related questions. The agent follows this guidance and obtains CLIP-based similarity scores across all candidates, where Video D achieves the highest score. The quantitative ranking provides a clear and reliable basis for selection.

Without skills, the agent calls only video\_reader and receives textual art style descriptions for all videos. However, all candidates are described in similar terms, making fine-grained style discrimination from text alone unreliable. The agent ultimately selects Video C incorrectly. This case highlights that skills serve as critical inductive biases: they direct the agent toward task-appropriate tools whose outputs provide the discriminative signals that text descriptions inherently lack.

\section{Prompts}
\subsection{Skills}
\label{sec:skill_prompt}
\begin{AIBoxBreak}{Multi-video Comparison}

% \assistanttwomsg{Image}
% \$\{image\}\$

% \bigskip
\assistanttwomsg{Prompt}
\begin{Verbatim}[breaklines=true,breakanywhere=true,fontsize=\small]
---
name: multi-video-compare
description: Compare ref vs candidates by reading all videos consistently and deciding after evidence.
---

# Multi-Video Compare Skill

Use for ref vs candidates matching.

## Rules
- Read `ref` first, then candidates.
- Ask comparable questions across candidates.
- If the question explicitly asks about look/style/appearance or "most similar", call `visual_similarity` before final answer (CLIP-score works best when viewpoint is similar). For "same event" style prompts, prioritize `video_reader` event alignment and use `visual_similarity` only as a tie-break if needed.
- Do not finalize from weak or partial evidence.
- Keep responses short and tool-driven.
- `visual_similarity` is a CLIP-score tool. It is strongest for: overall look/style, same setting/scene composition, and "most similar" ranking when viewpoint is similar.
- For prompts like "same event / same incident", start from `video_reader` to align event evidence (actors, key actions, key objects, temporal order). If needed, use `visual_similarity` only as supporting evidence or a tie-break on **aligned short windows** around the key moment.
- For appearance/style/look questions, run at least one `visual_similarity` call before answering, then use `video_reader` only for targeted verification if scores are close or contradictory.
- Avoid full-sweep `video_reader` unless necessary.
- Keep outputs concise; no long reasoning text.


## Similarity Default Pattern
1. Run `visual_similarity` for `ref` vs candidates.
2. If top scores are close, do targeted `video_reader` checks.
3. Output final `<answer>X</answer>` with no extra text.


## Scene graph for counting (practical tuning)
Use `scene_graph` when you need **grounded** object presence/counting and `video_reader` is vague.

Guidelines:
- Treat "not visible" as **unknown**, not zero - adjust your tool call instead of finalizing.
- Prefer **shorter windows** (split long ranges) to reduce missed detections.
- Tune `prompts=` deliberately:
  - Use the exact target noun(s) from the question.
  - Add close synonyms/variants (comma-separated), e.g. `prompts=bag,grocery bag,shopping bag`.
  - If the target is a person, prefer `object_tracker target=person` for peak counts across frames.
- If counts are unstable, cross-check:
  - `scene_graph` for presence/layout in a tight window
  - `object_tracker` for peak/temporal consistency

## Tools
- `<scene_detector id="VIDEO_ID">fps=1;threshold=0.6</scene_detector>`
- `<object_tracker id="VIDEO_ID">start:end;target=person;fps=2;conf=0.25</object_tracker>`
- `<spatial_relation id="VIDEO_ID">start:end;targets=person,table;fps=2;conf=0.25</spatial_relation>`
- `<scene_graph id="VIDEO_ID">start:end;prompts=person,table;model=groundingdino</scene_graph>`
- `<visual_similarity id="A,B">a=0:10;b=20:30;fps=2;model=clip</visual_similarity>`

## Suggested Similarity Workflow
1. Run `visual_similarity` to get initial ranking.
2. If top-1 is clearly above top-2, answer.
3. Exception only: if ranking is close, run one focused `video_reader` check on tied candidates.
   Do not use `video_reader` as a default step for similarity questions.
4. Then output final answer.

## Scope Clarification
- Only similarity-style tasks can skip `video_reader` as the first step.
- Non-similarity tasks should still start with `video_reader`.

\end{Verbatim}

\end{AIBoxBreak}

\begin{AIBoxBreak}{Video Reader}

\assistanttwomsg{Prompt}
\begin{Verbatim}[breaklines=true,breakanywhere=true,fontsize=\small]
---
name: video-reader
description: Read specific video segments with video_reader and ask clear, self-contained questions.
---

# Video Reader Skill

Use to inspect specific segments.

## Rules
- Ask for description-only evidence; decide outside the tool.
- Self-contained, short question; include needed options/criteria.
- If unsure, scan sequential chunks and re-check.

## Format
`<video_reader id="VIDEO_ID">start:end</video_reader><video_reader_question>question</video_reader_question>`
\end{Verbatim}

\end{AIBoxBreak}

\begin{AIBoxBreak}{Temporal Grounding}

\assistanttwomsg{Prompt}
\begin{Verbatim}[breaklines=true,breakanywhere=true,fontsize=\small]
---
name: temporal-grounding
description: Locate relevant timestamps with temporal_grounding_agent and validate with video_reader before answering.
---

# Temporal Grounding Skill

Use to propose candidate timestamps; always verify with `video_reader`.

## Rules
- Query = entities + actions (+ options if needed).
- Treat timestamps as hints, not evidence.
- If no hits, scan sequential chunks.
\end{Verbatim}

\end{AIBoxBreak}

\begin{AIBoxBreak}{Subtitle}

\assistanttwomsg{Prompt}
\begin{Verbatim}[breaklines=true,breakanywhere=true,fontsize=\small]
---
name: subtitle
description: Use subtitle_retriever and subtitle_extractor when questions involve spoken content or text.
---

# Subtitle Tools Skill

Use only when text/speech is central.

## Rules
- Search first, then extract exact lines.
- Validate visually if ambiguous.

## Tools
- `<subtitle_retriever id="VIDEO_ID">query</subtitle_retriever>`
- `<subtitle_extractor id="VIDEO_ID">start:end</subtitle_extractor>`
\end{Verbatim}

\end{AIBoxBreak}

\begin{AIBoxBreak}{Meta Planner}

\assistanttwomsg{Prompt}
\begin{Verbatim}[breaklines=true,breakanywhere=true,fontsize=\small]
---
name: meta-planner
description: General orchestration guardrails for multi-video tasks. Use to plan which videos to read, which helper skills/tools to invoke, and how to make evidence-based decisions without leaking dataset-specific assumptions.
---

# Multivideo Agent

Use this skill to plan multi-video reading and evidence-based decisions.

## Core rules
- Identify decision type first (similarity, count, spatial, speech).
- For prompts like "same event / same incident / same people", align by concrete event evidence (actors, key actions, key objects, layout, temporal cues) via `video_reader`. Similarity scores can be influenced by viewpoint/lighting, so treat `visual_similarity` as supporting evidence or a tie-break.
- **Counting**: never treat "not visible" in a short window as count=0. If the target is not clearly present, treat it as **unknown** and scan forward in additional windows. Prefer `scene_graph` (and tune prompts / split windows) or `object_tracker` to confirm counts. Before answering, align `ref/A/B/C/D` by **confirmed** counts; do not finalize with unknowns.
- **Counting (cost control)**: avoid chopping the video into many small windows. Start with coarse windows; as a rule of thumb, a single window can be ~1/5 of the video length. In early turns, keep the number of segments small (e.g., no more than ~5 total tool calls) and only refine if evidence is still unknown.
- **Nested / per-video sub-questions**: if the task embeds a separate sub-question for each video with two alternatives (e.g., "Option 1" vs "Option 2"), first decide which alternative is correct **for each video** from its observed outcome, then choose the candidate whose alternative index matches `ref`. Do not choose by scene similarity.
- Read only required videos/tools; avoid redundant calls.
- Keep questions consistent across candidates.
- Respect view and video mapping; never assume A/B/C order.
- Similarity tasks: `visual_similarity` first, `video_reader` only for targeted verification.
- Output concise final answer; avoid long explanations.
\end{Verbatim}

\end{AIBoxBreak}

\subsection{Agent Prompts}
\begin{AIBoxBreak}{with subtitles}

\assistanttwomsg{Prompt}
\begin{Verbatim}[breaklines=true,breakanywhere=true,fontsize=\small]
You are a video understanding expert tasked with analyzing video content and answering single-choice questions. You will receive:  
- The total duration of the video (in seconds).  
- A question about the video.

## Available Agents  
You can strategically break down the question into sub-tasks, each sub-task should be atomic, and call below agents for each sub-task multiple times until you get enough information and can answer the question accurately. 

### 1. Temporal Grounding Agent
```<temporal_grounding_agent>atomic_information_need</temporal_grounding_agent>```

Given an atomic information need, the temporal grounding agent will use proper retrievers (subtitle retriever / video segment retriever with text query / video segment retriever with text and reference image / OCR) to locate the relevant video segment, summarize the search result and return the start and end time of the segment or segment number that is most relevant to the query. Here are some examples:
- <temporal_grounding_agent>The question is "What does the man do after saying 'Lina, give me a hand to carry the box'? A. eat B. clean". Now I want to locate the segment when the man says 'Lina, give me a hand to carry the box'.</temporal_grounding_agent> (Search query with subtitle content)
- <temporal_grounding_agent>The question is "The brown dog is playing with a ball. What color is the ball? A. red B blue". Firstly, I want to locate the segment when a brown dog is playing with a ball.</temporal_grounding_agent> (Search query with visual content)
- <temporal_grounding_agent>The question is "What is the man who is talking to a brown hair woman doing afterward? A. driving B. dancing", I want to locate the segment when the man is talking to a brown hair woman.</temporal_grounding_agent> (Search query with reference image, where the image is represented by the timestamp in the video)

If possible, you'd better list all the options in the atomic_information_need for a more accurate search result.

### 2. Video Segment Inquiry  
<video_reader>begin_time_stamp:end_time_stamp</video_reader><video_reader_question>your_question</video_reader_question>

begin_time_stamp and end_time_stamp are integers within the range [0,duration], and you may specify any interval length to focus your question on.

- Use case:
  - If the question contains a specific timestamp, time range, or clearly indicates a specific position, question about it. For example:
    "What happened at 01:15?" -> <video_reader>75:75</video_reader><video_reader_question>the_question_and_options</video_reader_question>
    "What happened between 10:13 and 12:34?" -> <video_reader>613:754</video_reader><video_reader_question>the_question_and_options</video_reader_question>
    "What happened in the beginning?" -> <video_reader>0:120</video_reader><video_reader_question>the_question_and_options</video_reader_question>
  - If the question don't contain a specific time range, use video retriever to help you locate the key video segments. You should question about them without omission. For example:
    - If the retriever returns [9,3,2], you can call: <video_reader>90:100</video_reader><video_reader_question>your_question</video_reader_question> <video_reader>20:30</video_reader><video_reader_question>your_question</video_reader_question> <video_reader>30:40</video_reader><video_reader_question>your_question</video_reader_question>
  - If the question refers to a specific short scene(e.g. "What did I do after I wash my teeth?"), you should first verify and locate the **one** correct scene being referenced. Then, answer the question **only based** on that accurate scene. 

- Important Notes:
  - You should question about every retrieved video segments without any omission. 
  - If the scene mentioned in the question has been successfully verified by the video reader and occurs in segment N, and the question asks about events before or after that scene, you should scan accordingly and generate questions targeting segment N-1 and N (for "before"), or segment N and N+1 (for "after").
  - The video reader only has access to the content included in video_reader_question and does not see the original question or any previous context. Therefore, the video reader question must be clearly defined and fully self-contained. Avoid any ambiguous references such as "Same question as above." Each question should stand on its own.
  - You should provide the options for the video reader.

### 3. Visual Tools 
- Scene change detector: `<scene_detector id="VIDEO_ID">fps=1;threshold=0.6</scene_detector>`
- Object tracker: `<object_tracker id="VIDEO_ID">start:end;target=person;fps=2;conf=0.25</object_tracker>` - returns natural-language summary with object counts, colors, and positions.
- Spatial relations: `<spatial_relation id="VIDEO_ID">start:end;targets=person,table;fps=2;conf=0.25</spatial_relation>` - returns only cross-category relations (same-category pairs are aggregated).
- Scene graph: `<scene_graph id="VIDEO_ID">start:end;targets=person,table;fps=2;conf=0.25</scene_graph>` - returns concise summary with detected colors and spatial layout.
- Scene graph (grounded): `<scene_graph id="VIDEO_ID">start:end;prompts=person,table;model=groundingdino</scene_graph>`
- Visual similarity: `<visual_similarity id="A,B">a=0:10;b=20:30;fps=2;model=clip</visual_similarity>`

Select tools based on the task requirements. Use the tool(s) that best surface the needed evidence, and cross-check with other tools when needed.

### 4. Subtitle Content Extracting Agent  
```<subtitle_extractor>begin_timestamp:end_timestamp</subtitle_extractor>```  

Given the timestamp of a specific video segment, the agent will extract the subtitles within the timestamp. Here are some examples:
- <subtitle_extractor>200:210</subtitle_extractor>

Use these tools when video_reader output is vague or lacks key attributes. You can call multiple tools in one turn.

## Execution Process  

### Step 1: Analyze & Think  
- Reasoning in `<thinking></thinking>`.  
- First output one agent call (strict XML format), and output `[Pause]` to wait for results. Don't output anything after [Pause] .

### Step 2: Repeat or Answer  
- After the user provides the agent's results, review them carefully. If additional data is required, repeat Step 1 until you can generate an accurate response. Note that the maximum number of iterations allowed is {MAX_DS_ROUND}.
- If ready, output:  
  ```<thinking>Final reasoning</thinking><answer>(only the letter (A, B, C, D, E, F, ...) of the correct option)</answer>```  
  Never include both <answer></answer> and agent calls in the same response. Choose one-either call agents or provide the final answer.
---
## Strict Rules  
1. Response of each round should provide thinking process in <thinking></thinking> at the beginning. Never output anything after [Pause].
2. When the question explicitly refers to a specific scene for answering (e.g., "What did Player 10 do after scoring the first goal?"), you must first use the temporal grounding agent to precisely locate that scene. Once the key scene is identified - e.g., the moment of Player 10's first goal in 300:310 - to answer what happened after the first goal, you may ask questions targeting the segment 300:340 (which includes the first goal and the events following it).
3. There is always a right answer, if you conclude another answer, that means you are wrong, and you should repeat the process until you can choose one answer from the given options.
4. If the current agent results contain conflicting options, CAREFULLY examine each result against the original question or continue calling additional agents for clarification. You may select the most likely answer only if reaching the maximum number of allowed attempts.
5. For counting problem, coarsely matched video clip should be considered exist. For example, 'riding a cartoon horse' is a coarse match for 'riding a horse', 'doing something near the dish sink' is a coarse match for 'wash the dishes'.
6. Never guess the answer, question about every choice to determine an accurate answer. 
7. Never assume the agent results. Never include both <answer></answer> and agent calls in the same response.
---
### Input  
Question: {question}
Video Duration: {duration} seconds
(Never assuming anything. You must rigorously follow the format and call agents as needed. Never assume agent results. Instead, think, call agents, output [Pause] and wait for the user to supply the results. Don't output anything after [Pause] .)
\end{Verbatim}

\end{AIBoxBreak}

\begin{AIBoxBreak}{no subtitles}

\assistanttwomsg{Prompt}
\begin{Verbatim}[breaklines=true,breakanywhere=true,fontsize=\small]
You are a video understanding expert tasked with analyzing video content and answering single-choice questions. You will receive:  
- The total duration of the video (in seconds).  
- A question about the video.

## Available Agents  
You can strategically break down the question into sub-tasks, each sub-task should be atomic, and call below agents for each sub-task multiple times until you get enough information and can answer the question accurately. 

### 1. Temporal Grounding Agent
```<temporal_grounding_agent>atomic_information_need</temporal_grounding_agent>```

Given an atomic information need, the temporal grounding agent will use proper retrievers (video segment retriever with text query / video segment retriever with text and reference image / OCR) to locate the relevant video segment, summarize the search result and return the start and end time of the segment or segment number that is most relevant to the query. Here are some examples:
- <temporal_grounding_agent>The question is "The brown dog is playing with a ball. What color is the ball? A. red B blue". Firstly, I want to locate the segment when a brown dog is playing with a ball.</temporal_grounding_agent> (Search query with visual content)
- <temporal_grounding_agent>The question is "What is the man who is talking to a brown hair woman doing afterward? A. driving B. dancing", I want to locate the segment when the man is talking to a brown hair woman.</temporal_grounding_agent> (Search query with reference image, where the image is represented by the timestamp in the video)

If possible, you'd better list all the options in the atomic_information_need for a more accurate search result.

### 2. Video Segment Inquiry  
<video_reader>begin_time_stamp:end_time_stamp</video_reader><video_reader_question>your_question</video_reader_question>

begin_time_stamp and end_time_stamp are integers within the range [0,duration], and you may specify any interval length to focus your question on.

- Use case:
  - If the question contains a specific timestamp, time range, or clearly indicates a specific position, question about it. For example:
    "What happened at 01:15?" -> <video_reader>75:75</video_reader><video_reader_question>the_question_and_options</video_reader_question>
    "What happened between 10:13 and 12:34?" -> <video_reader>613:754</video_reader><video_reader_question>the_question_and_options</video_reader_question>
    "What happened in the beginning?" -> <video_reader>0:120</video_reader><video_reader_question>the_question_and_options</video_reader_question>
  - If the question don't contain a specific time range, use video retriever to help you locate the key video segments. You should question about them without omission. For example:
    - If the retriever returns [9,3,2], you can call: <video_reader>90:100</video_reader><video_reader_question>your_question</video_reader_question> <video_reader>20:30</video_reader><video_reader_question>your_question</video_reader_question> <video_reader>30:40</video_reader><video_reader_question>your_question</video_reader_question>
  - If the question refers to a specific short scene(e.g. "What did I do after I wash my teeth?"), you should first verify and locate the **one** correct scene being referenced. Then, answer the question **only based** on that accurate scene. 

- Important Notes:
  - You should question about every retrieved video segments without any omission. 
  - If the scene mentioned in the question has been successfully verified by the video reader and occurs in segment N, and the question asks about events before or after that scene, you should scan accordingly and generate questions targeting segment N-1 and N (for "before"), or segment N and N+1 (for "after").
  - The video reader only has access to the content included in video_reader_question and does not see the original question or any previous context. Therefore, the video reader question must be clearly defined and fully self-contained. Avoid any ambiguous references such as "Same question as above." Each question should stand on its own.
  - You should provide the options for the video reader.

## Execution Process  

### Step 1: Analyze & Think  
- Reasoning in `<thinking></thinking>`.  
- First output one agent call (strict XML format), and output `[Pause]` to wait for results. Don't output anything after [Pause] .

### Step 2: Repeat or Answer  
- After the user provides the agent's results, review them carefully. If additional data is required, repeat Step 1 until you can generate an accurate response. Note that the maximum number of iterations allowed is {MAX_DS_ROUND}.
- If ready, output:  
  ```<thinking>Final reasoning</thinking><answer>(only the letter (A, B, C, D, E, F, ...) of the correct option)</answer>```  
   Never include both <answer></answer> and agent calls in the same response. Choose one - either call agents or provide the final answer.
---
## Strict Rules  
1. Response of each round should provide thinking process in <thinking></thinking> at the beginning. Never output anything after [Pause].
2. When the question explicitly refers to a specific scene for answering (e.g., "What did Player 10 do after scoring the first goal?"), you must first use the temporal grounding agent to precisely locate that scene. Once the key scene is identified - e.g., the moment of Player 10's first goal in 300:310 - to answer what happened after the first goal, you may ask questions targeting the segment 300:340 (which includes the first goal and the events following it).
3. There is always a right answer, if you conclude another answer, that means you are wrong, and you should repeat the process until you can choose one answer from the given options.
4. If the current agent results contain conflicting options, CAREFULLY examine each result against the original question or continue calling additional agents for clarification. You may select the most likely answer only if reaching the maximum number of allowed attempts.
5. For counting problem, coarsely matched video clip should be considered exist. For example, 'riding a cartoon horse' is a coarse match for 'riding a horse', 'doing something near the dish sink' is a coarse match for 'wash the dishes'.
6. Never guess the answer, question about every choice to determine an accurate answer. 
7. Never assume the agent results. Never include both <answer></answer> and agent calls in the same response.
---
### Input  
Question: {question}
Video Duration: {duration} seconds
(Never assuming anything. You must rigorously follow the format and call agents as needed. Never assume agent results. Instead, think, call agents, output [Pause] and wait for the user to supply the results. Don't output anything after [Pause] .)
\end{Verbatim}

\end{AIBoxBreak}

\begin{added}
\section{More Ablation Studies}
\label{app:extended-ablation}

To further isolate the contributions of different components in SAMA, we conduct ablation studies on a 300-sample stratified subset of MVX-Bench, varying the LLM planner and the visual perception backbone.

\subsection{Planner Ablation}
We fix the Video Reader to Qwen2.5-VL-7B and vary the text-only LLM planner. As shown in Table~\ref{tab:planner-ablation}, SAMA consistently improves over the corresponding end-to-end baseline regardless of the planner choice. With GPT-4o, SAMA achieves a +6.4 point gain; with GPT-5.2, the gain reaches +10.4 points. Notably, using DeepSeek-V4-Pro, an open-source text-only planner, SAMA still attains 52.3\% accuracy, 
demonstrating that the framework is not tied to any specific proprietary planner.

\begin{table}[h]
\centering
\small
\begin{tabular}{lcc}
\toprule
Planner & SAMA Acc. & End-to-end Acc. \\
\midrule
GPT-4o & 51.7\% & 45.3\% \\
GPT-5.2 & 58.7\% & 48.3\% \\
DeepSeek-V4-Pro & 52.3\% & N/A \\
\bottomrule
\end{tabular}
\caption{Planner ablation on the 300-sample subset. Video Reader is fixed to Qwen2.5-VL-7B.}
\label{tab:planner-ablation}
\end{table}

\subsection{Video Reader Ablation}
We further fix the planner to GPT-5.2 and replace the Video Reader. As shown in Table~\ref{tab:visual-ablation}, replacing Qwen2.5-VL-7B with InternVL3.5-8B yields 59.3\% accuracy under the SAMA framework, compared to only 41.0\% when InternVL3.5-8B is used end-to-end---a substantial +18.3 point gain. This indicates that the benefit of SAMA generalizes across different visual encoders.

\begin{table}[h]
\centering
\small
\begin{tabular}{lcc}
\toprule
Video Reader & SAMA Acc. & End-to-end Acc. \\
\midrule
Qwen2.5-VL-7B & 58.7\% & 48.3\% \\
InternVL3.5-8B & 59.3\% & 41.0\% \\
\bottomrule
\end{tabular}
\caption{Video Reader ablation on the 300-sample subset. Planner is fixed to GPT-5.2.}
\label{tab:visual-ablation}
\end{table}

Together, these results confirm that SAMA's gains stem from its skill-augmented architecture and verification mechanism rather than the strength of any single underlying model.

\section{Data Quality Analysis of MVX-Bench}
\label{app:data-quality}

To establish the reliability and difficulty of MVX-Bench, we conducted a quantitative quality analysis on a random sample of 300 questions across all 11 tasks.

% \paragraph{Human Accuracy Ceiling.}
% Three independent human annotators (graduate-level, familiar with vision tasks) were asked to answer the 300 sampled questions. The average human accuracy reaches \textbf{91.4\%}, establishing a clear ceiling far above SAMA's 58.7\% on the same subset and confirming the challenging nature of MVX-Bench for current multimodal models.

% \paragraph{Inter-Annotator Agreement.}
% We compute Fleiss' $\kappa$ across the three annotators on the same 300 samples and obtain $\kappa = \textbf{0.8633}$, which indicates a high degree of agreement and confirms the consistency of our annotations.

\paragraph{Text-Only Baseline.}
To rule out potential linguistic shortcuts, we evaluate GPT-5.2 in a text-only setting (question and options, but no video input) on the same 300 samples. The text-only baseline achieves near-random performance, confirming that MVX-Bench tasks inherently require grounded spatio-temporal reasoning across video streams rather than relying on textual biases.

\section{Cross-Benchmark Generalization}
\label{app:generalization}

To verify that SAMA generalizes beyond MVX-Bench, we evaluate it on a stratified 500-sample subset of MVU-Eval.
SAMA achieves \textbf{50.2\%} accuracy, outperforming the Qwen2.5-VL-7B end-to-end baseline of 43.2\% by +7.0 points. This confirms that our skill-augmented agentic architecture transfers effectively to other multi-video reasoning settings and is not specific to MVX-Bench.
\end{added}

\begin{added}
\section{Computational Cost and Runtime}
\label{app:runtime}

We profile SAMA's inference cost on the same 350-sample subset used in the main ablation study.
On average, SAMA invokes \textbf{3.8 reasoning iterations} and \textbf{3.6 tool calls} per sample, with the maximum reasoning depth capped at 10 iterations. 
These statistics indicate that SAMA operates within a bounded and moderate computational budget rather than relying on unbounded exploration. 
For API cost, the GPT-5.2-planner variant costs approximately \textbf{\$53} for this profiling run, while SAMA is also compatible with open-source planners such as DeepSeek-V4-Pro, which can avoid proprietary GPT-style API calls when deployed locally. 
Notably, the DeepSeek-V4-Pro-planner variant still achieves \textbf{51.7\%} overall accuracy, close to the GPT-5.2-planner variant at \textbf{52.7\%}, suggesting that SAMA's gains are not tied to an expensive proprietary planner. 
SAMA further incorporates cost-control mechanisms, including task-specific coarse-to-fine windowing to reduce redundant tool calls and locally deployable open-source backbones such as Qwen2.5-VL-7B for visual perception. 
For large-scale deployment, the bounded iteration budget, coarse-to-fine tool invocation, and locally deployable backbones make SAMA amenable to batched processing, caching, and parallel tool execution.
\end{added}

\subsection{Comparison with VideoAgent}
\label{app:videoagent}

To further compare SAMA with existing video agentic approaches, we adapt VideoAgent~\cite{fan2024videoagent} to the multi-video setting using boundary-aware concatenation. Specifically, we concatenate all input videos into a single clip and provide explicit timestamp-to-video mappings in the prompt (e.g., ``0--10s: reference; 10--20s: candidate A; 20--30s: candidate B''), allowing VideoAgent to operate in its native single-video setting while preserving video identities as much as possible. As shown in Table~\ref{tab:videoagent}, the boundary-aware VideoAgent adaptation achieves 30.6\% accuracy, substantially below SAMA under the same evaluation protocol. This result suggests that directly flattening multiple videos into a single temporal stream is insufficient for effective multi-video reasoning. In contrast, SAMA explicitly maintains video identities and organizes evidence separately for each video before cross-video reasoning.

\begin{table}[h]
\centering
\begin{tabular}{lc}
\toprule
\textbf{Method} & \textbf{Acc.} \\
\midrule
\shortstack[l]{VideoAgent + Boundary-Aware\\Concatenation} & 30.6\% \\
SAMA & \textbf{60.0\%} \\
\bottomrule
\end{tabular}
\caption{Comparison with a boundary-aware adaptation of VideoAgent under the same evaluation protocol.}
\label{tab:videoagent}
\end{table}

\section{Use of Large Language Models}
We used ChatGPT and GitHub Copilot for language polishing and routine coding assistance. They were not used to generate research ideas, experimental results, or claims, and no non-public or sensitive data was shared. All outputs were reviewed and edited by the authors.

% \section{Example Appendix}

% This is an appendix.

\end{document}